\documentclass{article}

\usepackage[T1]{fontenc}
\usepackage{microtype}
\usepackage{graphicx}
\usepackage{pgfplots}
\pgfplotsset{compat=1.16}
\usepackage{booktabs}
\usepackage{hyperref}
\usepackage{xurl}

\usepackage[preprint]{icml2026}
\hypersetup{pdftitle={Spec2Weights: Certifying Networks that a Language Model Writes from a Specification},
  pdfsubject={arXiv preprint}}

\usepackage{amsmath}
\usepackage{amssymb}
\usepackage{mathtools}
\usepackage{amsthm}
\usepackage[capitalize,noabbrev]{cleveref}
\usepackage{xcolor}

\theoremstyle{plain}
\newtheorem{theorem}{Theorem}
\newtheorem{proposition}{Proposition}
\newtheorem{lemma}{Lemma}

\theoremstyle{definition}

\theoremstyle{remark}
\newtheorem{remark}{Remark}
\crefname{theorem}{Theorem}{Theorems}
\crefname{proposition}{Proposition}{Propositions}
\crefname{lemma}{Lemma}{Lemmas}
\crefname{remark}{Remark}{Remarks}

\newcommand{\Z}{\mathbb{Z}}

\icmltitlerunning{Spec2Weights: Certifying Networks a Language Model Writes}

\begin{document}

\twocolumn[
\icmltitle{Spec2Weights: Certifying Networks that a Language Model\texorpdfstring{\\}{ }Writes from a Specification}

\begin{icmlauthorlist}
\icmlauthor{Soumik Bhattacharjee}{buet}
\icmlauthor{Musa Tur Farazi}{buet}
\icmlauthor{Md Nasir Uddin}{buet}
\icmlauthor{Tishna Sabrina}{uap}
\icmlauthor{Chowdhury Mohammad Rakin Haider}{buet}
\end{icmlauthorlist}

\icmlaffiliation{buet}{Department of Computer Science and Engineering, Bangladesh University of Engineering and Technology (BUET), Dhaka, Bangladesh}
\icmlaffiliation{uap}{Department of Electrical and Electronic Engineering, University of Asia Pacific (UAP), Dhaka, Bangladesh}
\icmlcorrespondingauthor{Soumik Bhattacharjee}{2005031@cse.buet.ac.bd, soumikbhat330@gmail.com}

\icmlkeywords{weight-space programming, verification, compiled networks, floating point, statistical query}

\vskip 0.3in
]

\printAffiliationsAndNotice{}

\begin{abstract}
A language model can write the weights of a small MLP, CNN, RNN or GNN from a plain-language specification, with
no data and no gradient step. We ask when such a network can be \emph{proved} exact in the float32 arithmetic it
actually runs in. A float-safety lemma makes exhaustive float32 enumeration a proof for any summation order, and a
layered error bound covers networks whose weights are not short binary fractions. With dynamic programming,
template checks, an SMT proof, breadth-first search and induction for weight-tied recurrences, this
certifies 12 of 15 written networks (one after a single constant fix) on every input of a stated domain (up to $10^{16}$ inputs, or every graph or grid
up to a size bound), in a minute of CPU or less each. The certificates also say where a network stops being
exact: the shortest-path network is exact iff the distance is at most 4,096 and fails at the
predicted hops. Thirty new specifications were each written by a fresh session of one model family (Claude Opus
5.5). Twenty-five first submissions are certified exact for any summation order; the five tax schedules stay within
\$0.11 of the true tax at every whole dollar in one fixed float32 run. An open 27B model passes 23 of the 30, and 25 with
more thinking and code execution, so the method transfers in part. We also prove two limits of learning from
examples: input coordinates never active in training cannot be learned from more in-range data, and Luhn-type
checksums have exponentially small target-dependent gradient variance across their family. Tuned gradient descent on the same
architectures agrees with both, which cover two of our five failure classes.
\end{abstract}


\section{Introduction}
\label{sec:intro}

Large language models can write the weights of small networks directly. This is called weight-space programming
(WSP) \citep{takehi2026sparks}. Given the specification of a card-number checksum, an arithmetic-logic unit or a
routing table, the model writes a short Python program that sets every weight of a fixed MLP, CNN, RNN or GNN. No
data and no gradient step are used, and the result is often exact. For such a network to be used in place of
tested code, two questions matter. Can exactness be \emph{proved}, in the arithmetic the network actually runs in,
and not only tested on samples? And where does the network stop being exact?

We answer both for a suite of 15 written networks and 30 new specifications. We also compare with gradient descent
(GD) on the same architectures, and prove two limits of learning these functions from examples.
\begin{itemize}
\item \textbf{Certificates for the float32 network.} A float-safety lemma (\cref{thm:t1}) makes exhaustive float32
enumeration a proof for any summation order. A layered error bound (\cref{lem:margin}) covers weights that are not
short binary fractions. With dynamic programming, template checks, an SMT proof, a breadth-first search and a local
induction for weight-tied recurrences (\cref{thm:t2}), we certify 12 of the 15 networks on stated domains of up to
$10^{16}$ inputs, or on every graph or grid up to a size bound (\cref{tab:suite,tab:envelope}). Every certificate is
carried to float32 by \cref{thm:t1} or \cref{lem:margin} (App.~\ref{app:float}).
\item \textbf{Where networks break.} The certificates say where each network stops being exact. The shortest-path
network is exact iff the distance is at most $B=4096$. So it first fails at hop 820 of a path with edge length 5,
and at hop 4,097 of a unit path; both were predicted before the run and held. Other predictions were looser bands
(\cref{tab:x3}).
\item \textbf{New specifications.} Thirty specifications made by a seeded generator were each written by a fresh
session of one model family. Twenty-five first submissions are certified exact; the five tax schedules pass a
test-based rule fixed in advance (\cref{sec:x6}).\item \textbf{Two limits of learning from examples.} An input coordinate never active in training cannot be learned
from more in-range data (\cref{thm:t3}). Luhn-type checksums have exponentially small target-dependent gradient variance across
their family (\cref{thm:t4}). Tuned GD on day of week (with one-hot and with scalar century inputs) and on Luhn (with the
training mixture of the suite and with uniform digits) is consistent with both. These two results cover two of the
five failure classes we see. The other three classes have no
theorem here.
\end{itemize}
The writer is a language model that authors a weight-building Python program. That is close to program-to-weights
compilation; what differs is that the source is plain language, not a DSL (\cref{sec:related}). A certified
written network is useful where a network must be shipped: a fixed accelerator, a model mixing learned and exact
parts, or a test bed with a known answer. For a stand-alone rule, the program itself is simpler.


\section{Related work and what is new}
\label{sec:related}

\paragraph{Rules compiled into networks.} Knowledge-based networks \citep{towell1994kbann} compiled rules into
initial weights and then trained them. Exact recurrent constructions go back to \citet{siegelmann1995computational}.
For transformers, RASP \citep{weiss2021thinking} is a language whose programs Tracr \citep{lindner2023tracr} compiles
into weights. ALTA \citep{shaw2024alta} adds loops and studies cases where a solution can be expressed but training
does not find it; this is the closest prior work to our GD comparison. TracrBench \citep{thurnherr2024tracrbench}
uses an LLM to write RASP programs, which Tracr then compiles. Compiled testbeds such as InterpBench
\citep{gupta2024interpbench}, learned transformer programs \citep{friedman2023learning}, looped transformers
\citep{giannou2023looped} and $n$-gram constructions \citep{svete2024ngram} are further lines. Other compilers produce
modules inside LMs \citep{weng2023mastering}, learned surrogates of programs \citep{weber2024learning}, frozen modules
from Scheme-like programs (the Neural Compiler \citep{sheneman2026neural}, which is exact to floating point and
compares with PINN baselines, but not with GD on the same architecture) and fuzzy functions \citep{zhang2026paw}. Our
writers also author a program that builds the weights, but in Python from a plain-language specification, with no
compiler or DSL.

\paragraph{Weight-space programming.} The WSP blog \citep{takehi2026sparks} showed LLMs writing policy, generator and
language-model weights and reported empirical scores. A public repository \citep{trimm2026safetynet} has an LLM write
exact US benefits formulas against trained networks of the same architecture. It also reports GPT and Gemini writing
the same exact circuits, a cross-family control we lack for our 15-network suite. Our tax task
is therefore only a certificate example. iWSP \citep{mehta2026iwsp} builds interpretable models by programming
weights iteratively, and \citet{ahmed2026audit} audits LLM weight reading. Three concurrent works (November 2026)
are close to ours. WeightBench \citep{weightbench2026} benchmarks reading, fixing and writing weights on 42 tasks.
\citet{gross2026} has LLMs write policy weights and verifies the policies with a probabilistic model checker.
\citet{lehmann2026} has LLMs compile game rules into compact agents and finds that training the same architectures
from scratch, or with PPO, does far worse. None of them proves float32 exactness on a whole input domain, or proves
why training fails.
\paragraph{Why GD fails.} Parity-like targets defeat gradient methods \citep{shalevshwartz2017failures,abbe2020poly},
SGD learns sparse parities near the SQ limit \citep{barak2022hidden}, and SQ complexity has a general characterisation
\citep{feldman2016general}. ReLU networks extrapolate linearly \citep{xu2021extrapolate}, and length generalisation is
fragile \citep{zhou2023algorithms}. Learned channel decoders \citep{nachmani2016learning,gruber2017deep} and
differentiable SAT for Sudoku \citep{wang2019satnet} are the domain baselines. GNNs can express and partly learn
Bellman--Ford-like procedures \citep{xu2020reason,velickovic2020neural,velickovic2022clrs}; discrete neural
algorithmic reasoning \citep{rodionov2025discrete} proves its trained models correct.

\paragraph{Verification.} ReLU networks are verified by SMT or bound propagation
\citep{katz2017reluplex,wang2021beta}, and binarized networks exactly by SAT \citep{narodytska2018verifying}.
Verifiers that ignore floating-point error can be fooled \citep{jia2021exploiting,zombori2021fooling}, also through the
order of operations in deployment \citep{szasz2025soundness}; sound ones over-approximate rounding \citep{singh2019abstract}. Our certificates avoid this by
proving that the float32 computation equals the exact one (\cref{thm:t1}) or by bounding the gap (\cref{lem:margin}).
Written networks make this cheap, because their weights are small integers or short binary fractions.

\begin{table}[t]
\caption{Closest work. \checkmark: present; --: absent; (\checkmark): partly. $^a$probabilistic model checking of
the policy in its environment. ``Partial'': theorems for two of five classes.}
\label{tab:novelty}
\centering\footnotesize
\setlength{\tabcolsep}{2pt}
\resizebox{\columnwidth}{!}{%
\begin{tabular}{@{}lccccc@{}}
\toprule
 & who makes & non-transf. & float-sound & GD, same & theory of \\
 & the weights & targets & certificate & architecture & GD failure \\
\midrule
Tracr / ALTA & compiler (DSL) & -- & -- & ALTA (\checkmark) & -- \\
TracrBench & LLM via RASP & -- & -- & -- & -- \\
CLRS / discrete NAR & training & GNN & -- & trained & -- \\
Neural Compiler & compiler (DSL) & \checkmark & tests & -- & -- \\
WSP blog & LLM & \checkmark & -- & RL / GD & -- \\
Benefits repo & LLM, 3 vendors & MLP & tests & \checkmark & -- \\
WeightBench & LLM & \checkmark & -- & -- & -- \\
Gross (2026) & LLM & policies & (\checkmark)$^a$ & -- & -- \\
Lehmann (2026) & LLM & \checkmark & -- & \checkmark & -- \\
\textbf{This paper} & LLM (Python) & MLP, CNN, RNN, GNN & \checkmark & \checkmark & partial \\
\bottomrule
\end{tabular}}
\end{table}

What we claim is the last row of \cref{tab:novelty}: an LLM that writes a weight-building program from a
plain-language specification, non-transformer targets, certificates that hold for the float32 network in any
summation order, tuned GD on the same architecture, and theorems for two of the five GD failure classes we observe.


\section{Suite, protocol and certificates}
\label{sec:setup}

\paragraph{Suite.} \Cref{tab:suite} lists the 15 written networks. Each was written earlier, in our lab notebook, by
one session of the same model family as the writers below. Each session got a plain-language specification and an
architecture fixed beforehand; it used no data and no gradient step. App.~\ref{app:origin} gives the inclusion rule
and the notebook's record. By the notebook's own tests, the first submission was exact for 14 of the 15. The day-of-week network was not: its one
free constant was written as 4 from memory instead of 5, so every date was off by one day. We report it as
\emph{exact after one fix}. Some notebook claims were wrong; App.~\ref{app:origin} lists every correction. The suite
has two tiers: \emph{exact} (a certificate on a stated domain) and \emph{optimal-statistic} (no exact answer; we
measure the gap to the Bayes rule).

\begin{table*}[t]
\caption{The certified suite. ``nz'' = non-zero parameters. Certificate: \emph{enum} = exhaustive enumeration;
\emph{DP} = dynamic programme over reachable hidden pre-activations; \emph{tmpl} = weight-template check plus a
counting argument; \emph{SMT} = z3 proof per king square (Int sort); \emph{BFS} = search over reachable RNN states;
\emph{local} = \cref{thm:t2}; \emph{PL} = exact rationals at all breakpoints. Every certificate reasons in exact
integers or rationals; the last column says how it is carried to the float32 network (any summation order): T1 =
\cref{thm:t1} with the per-layer $\max_iM_i$ of App.~\ref{app:float}; L1 = \cref{lem:margin}. Times are CPU
wall-clock on one or two threads. Domains: $^a$5 operations on all pairs of 8-bit operands; $^b$at most one piece per
square, exactly one white king, kings not adjacent; $^c$neutral molecules over C, N, O, F, S, Cl with integer bond
orders 1--3 and bond-order sum at most the valence at every atom (hydrogen counts only); $^d$integer edge lengths
1--5 and at least $|V|-1$ rounds; $^e$at least $\max d$ steps and at most $1.39$M steps ($T>\max d$ keeps the guard off at $d=T$); $^f$reading frame 0.}
\label{tab:suite}
\centering\footnotesize
\setlength{\tabcolsep}{2.6pt}
\resizebox{\textwidth}{!}{%
\begin{tabular}{@{}llrrllll@{}}
\toprule
task & architecture & weights & nz & certified domain & certificate $\to$ float32 & result & time \\
\midrule
8-bit ALU & MLP 21-96-64-8 & 8,840 & 626 & 327,680 inputs$^a$ & enum, T1 & exact & 4.5\,s \\
day of week & MLP 83-64-192-7 & 19,207 & 4,105 & all dates 1600--2399 & enum, T1 & exact after 1 fix & 8\,s \\
Roman numerals & conv-sum-linear & 513 & 98 & numerals 1--3,999 & enum, L1 & exact (margin $57\times$ bound) & 0.2\,s \\
tic-tac-toe & MLP 27-256-9 & 9,481 & 1,524 & all games, all tie-breaks & enum, T1 & 0 losses & $<$2\,s \\
routing & MLP 32-32-8 & 1,320 & 226 & all $2^{32}$ addresses & enum ($2^{20}$ patterns), T1 & exact & 23\,s \\
Luhn & MLP 160-192-2 & 31,298 & 21,360 & all $10^{16}$ strings & DP (145 states), T1 & exact & 0.1\,s \\
Sudoku validity & MLP 729-243-2 & 177,878 & 2,674 & all $9^{81}$ filled grids & tmpl, T1 & exact & $<$1\,s \\
tax, 2024 rate schedule & MLP 3-32-1 & 161 & 56 & income \$0--2M, 2 statuses & PL, L1 & within \$1.90 & $<$5\,s \\
chess check & conv $15{\times}15$-sum & 194,618 & 2,337 & legal-type boards$^b$ & SMT, T1 & exact & 63\,s \\
chemistry H count & 1-layer MPNN & 127 & 20 & molecules$^c$ & tmpl, T1 & exact & $<$1\,s \\
shortest paths & weight-tied MPNN & 177 & 11 & all graphs$^d$ & local, T1 & exact iff $d\le4096$ & 2\,s \\
maze distance & weight-tied conv RNN & 225 & 30 & all grids, all sizes$^e$ & local, T1 & exact & $<$1\,s \\
gene finder & conv-RNN-MLP & 6,889 & 587 & all DNA, all lengths$^f$ & BFS (3 states), T1 & exact & $<$1\,s \\
Hamming(7,4) & 7-16-softmax-4 & 196 & 176 & BPSK, AWGN & Bayes gap & $\approx$ block ML & MC \\
motion & conv-mean-linear & 457 & 36 & random-dot pairs & Bayes gap & $=$ Bayes & MC \\
\bottomrule
\end{tabular}}
\end{table*}

\begin{table*}[t]
\caption{Operating envelope of each certified network. All certificates need IEEE binary32 (or wider) with
round-to-nearest in every multiply and add. Settings that break this: TF32 matrix units, float16 or bfloat16 autocast,
fast-math reassociation into lower precision, and Winograd or FFT convolution. Outside its domain a network gives wrong answers silently. The guard is a cheap
caller-side check; for the saturating networks the same test can be one extra ReLU unit. $^\ast$The 25 certified nets of \cref{sec:x6}.}
\label{tab:envelope}
\centering\scriptsize
\setlength{\tabcolsep}{3pt}
\begin{tabular}{@{}p{3.2cm}p{5.0cm}p{4.8cm}p{3.3cm}@{}}
\toprule
network & preconditions beyond the input format & behaviour outside them & cheap guard \\
\midrule
ALU, routing, Luhn, tic-tac-toe; fresh$^\ast$ check digits, routing & none (the domain is every encodable input; legal positions for tic-tac-toe) & -- & one-hot / bit check \\
day of week; fresh$^\ast$ calendars & a real date in 1600--2399 & some weekday, silently & validate the date \\
Roman numerals & canonical numeral, value 1--3,999 & a value by the additive rule & canonical-form regex \\
Sudoku validity & every cell filled & incomplete grids read as invalid & count filled cells \\
tax, 2024 schedule & income \$0--2M; status one-hot & error grows; status gate fails above \$3.03M & flag income $>\$2$M \\
chess check & one piece per square, one white king, kings apart & possible false check & one-hot, king count, kings apart \\
chemistry H count & integer bond orders, sum $\le$ valence & over-valent atom reads 0 hydrogens & compare sum with valence \\
shortest paths & lengths 1--5; $T\ge|V|-1$; float32 & distances above 4,096 read 4,096 & output $=4096$ $\Rightarrow$ flag \\
maze distance & $T\ge\max d$; $T\le1.39$M; float32 & cells with $d>T$ read $T$; float16 wrong from $T=684$ & output $=T$ $\Rightarrow$ flag \\
gene finder & letters A, C, G, T; frame 0 & unknown letters read as ``not in a gene'' & alphabet check \\
\bottomrule
\end{tabular}
\end{table*}

\paragraph{Certificates and float32.} Every certificate reasons about an exact model: integers or rationals. The
network actually runs in float32. \Cref{thm:t1} says the two agree, in any summation order, when a per-unit sum of
absolute terms stays at most $2^{24}$ in units of that layer's grid. We report this sum for every layer of every
network (App.~\ref{app:float}): it holds for all rows marked T1, with the largest value $2.1\times10^6$ among fixed-domain networks. The Roman
and tax networks use weights such as $0.01$, which are not short binary fractions; for them \cref{lem:margin} bounds
the float32 error instead.

\paragraph{Predictions logged before each run.} Before every experiment we wrote its prediction into a local log
(\texttt{EXPERIMENTS\_p14.md}; the Qwen runs have their own logs), and appended the result afterwards without editing
the prediction. The log has no external timestamp, so this record is self-attested. Table~\ref{tab:predictions}
(App.~\ref{app:preds}) lists every prediction, says which follow from a theorem, and gives each outcome.

\paragraph{GD baselines.} For ten tasks we ran GD on the same architecture with a grid: learning rate
$\{3\times10^{-4},10^{-3},3\times10^{-3}\}$, width $\times1$ (the written architecture) or $\times4$, and weight decay
off or AdamW $10^{-2}$, plus a curriculum where one is natural. For each configuration one seed is trained and scored
on a validation split drawn from the \emph{training} distribution. Selection runs are shorter than final runs. The
selected configuration is then trained with three seeds and scored once on the test sets. App.~\ref{app:gd} lists the
budgets, every grid cell's validation score, the test sizes and binomial intervals.

\paragraph{Fresh specifications.} We wrote a seeded generator (seed 20261123) and made 30 new specifications on
2026-11-23: 10 check-digit schemes (random modulus 7--13, length 8--14, random multipliers and digit tables), 10
routing tables (16--45 nested routes), 5 hypothetical tax schedules (4--8 brackets; two with a phased-out credit) and
5 calendars (week length 5--10, random month lengths and leap rules). The writer's documented knowledge cutoff is June
2026, as stated in the model identity the harness gives each session. The specific tables are random, so they cannot
be in any training data; the families themselves are familiar. Each spec went to one fresh session of
\texttt{claude-opus-5-5}. The session read one prompt file (App.~\ref{app:prompt}) and wrote a Python program that
builds the weights. The rules allow Python arithmetic on the spec and forbid reading other files, evaluating the
network, fitting and search. The first saved submission is final and is scored once. The scoring rules, including the
tax rule of at most \$1 error at every whole dollar, were fixed in the scorer before any session ran.

\paragraph{What each session saw.} We read the 30 session logs. Besides its prompt, each session had the harness's
standard context: a system prompt, tool definitions, a list of available skills and agent types, the platform and
working directory, the date, the model identity, and the project's auto-memory index. The index has 14 one-line
entries about the research programme, such as ``WSP MNIST study'' and ``Paper portfolio''. No project instruction
file existed. No entry names a task, a construction or any weight. The same context was in all 30 sessions; we did
not rerun with a bare context. The generator, the prompt and the 30 reference solutions were written by the author
session that ran the study, before any writer session. The prompt gives ``a per-digit table or an offset'' as an
example of allowed arithmetic; this hints at the construction, and it was in all 30 prompts.


\section{Theory: when a check is a proof, and two limits of learning}
\label{sec:theory}

For each result we cite what is already known and say what we add. Proofs are in App.~\ref{app:proofs}.

\subsection{T1: finite checks are proofs, in any summation order}

Layer $\ell$ computes $z_i=\sum_jW_{ij}x_j+b_i$. Every float32 number is a binary fraction. Write its inputs on a grid
$x_j=\hat x_j2^{-e_{\ell-1}}$, its weights as $W_{ij}=\hat W_{ij}2^{-k_\ell}$ and its bias as
$b_i=\hat b_i2^{-e_\ell}$, with integers $\hat x,\hat W,\hat b$ and $e_\ell=e_{\ell-1}+k_\ell$. The outputs then lie
on the grid $2^{-e_\ell}\Z$, which is the next layer's input grid.
\begin{theorem}[Float-exact enumeration]\label{thm:t1}
Assume every multiply and add is an IEEE binary32 (or wider) operation with round-to-nearest and gradual underflow (no flush-to-zero). If for every unit $i$ of
layer $\ell$ and every input of a finite domain $\mathcal D$, $M_i:=\sum_j|\hat W_{ij}\hat x_j|+|\hat b_i|\le 2^{24}$,
and $-103\le e_\ell\le149$ (no overflow or loss below the subnormal grid), then every partial sum of $z_i$ is computed exactly, in every order and with or without fused multiply--add. If this
holds for every layer (ReLU, max and sum pooling are layers too), the float32 network equals the exact rational
network on $\mathcal D$, and exhaustive enumeration of $\mathcal D$ in float32 is a proof that does not depend on the
summation order.
\end{theorem}
The fact behind it is standard \citep{higham2002accuracy}; its use as a proof rule is ours. The assumption fails under TF32 matrix units, 16-bit autocast, fast-math reassociation into lower precision, and
Winograd or FFT convolution; \cref{tab:envelope} repeats this. When weights are not short binary fractions, as in the
Roman reader (value$/100$), $M_i$ is huge and \cref{thm:t1} does not apply. The standard error bound \citep{higham2002accuracy}, applied layer by layer, does.
\begin{lemma}[Layered margin bound]\label{lem:margin}
Let $z_\ell=W_\ell h_{\ell-1}+b_\ell$, $h_\ell=\mathrm{relu}(z_\ell)$, $h_0=x$, be evaluated in binary32 with
round-to-nearest in any order and without overflow or underflow, giving $\hat z_\ell,\hat h_\ell$. Here $W_\ell,b_\ell$ are the stored float32 weights. Let $n_\ell$ be the fan-in plus one, and
$\gamma_n=nu/(1-nu)$ with $u=2^{-24}$ and $n_\ell u<1$. Define, entrywise, $\epsilon_0=0$ and
\[\epsilon_\ell=|W_\ell|\,\epsilon_{\ell-1}+\gamma_{n_\ell}\bigl(|W_\ell|(|h_{\ell-1}|+\epsilon_{\ell-1})+|b_\ell|\bigr).\]
Then $|\hat z_\ell-z_\ell|\le\epsilon_\ell$ and $|\hat h_\ell-h_\ell|\le\epsilon_\ell$. A unit with $z+\epsilon\le0$
outputs exactly 0 in float32, so it can be left out of the next layer's count $n$. If the exact output clears a
decision threshold by more than $\epsilon_L$, the float32 decision equals the exact one in every summation order.
\end{lemma}

\subsection{T2: a size-free certificate for weight-tied distance recurrences}

Consider a graph $G=(V,E)$ of any size with integer edge lengths in $[1,L]$, a source $s$, and a weight-tied
recurrence that keeps a closeness $c_t(v)\in\Z\cap[0,B]$ and updates
\begin{equation}\label{eq:upd}
\begin{aligned}
c_{t+1}(v)&=F\bigl(c_t(v),\,m_t(v)\bigr),\\
m_t(v)&=\textstyle\max_{u\in N(v)}\bigl(c_t(u)-w_{uv}\bigr),
\end{aligned}
\end{equation}
with $m_t(v)=0$ for an isolated node and $c_0=B\,\mathbf 1[v=s]$. The written shortest-path network has exactly this
form; a linear-algebra check on its weights shows that its two used units enter only through $c=p+q$. The induction is
Bellman--Ford's \citep{bellman1958routing}; the local check, clamp law and failure sizes are ours.
\begin{theorem}[Local certificate, all sizes]\label{thm:t2}
If $F(c,m)=\max(c,m)$ for all integers $c\in[0,B]$ and $m\in[-L,B]$ (a finite check), then on every graph
$c_t(v)=\max(0,B-d_t(v))$, where $d_t$ is the shortest distance over paths of at most $t$ edges. With $T\ge|V|-1$
rounds the output $B-c_T(v)$ equals $\min(d(v),B)$: the network is exact on node $v$ iff $d(v)\le B$. On a path with
edge length $\ell$ the first wrong node is hop $\lfloor B/\ell\rfloor+1$. For the written network, which computes $F$ as $\mathrm{relu}(c)+\mathrm{relu}(m-c)$ with integer weights,
$2B+L\le2^{24}$ makes the float32 run equal the integer one (\cref{thm:t1}).
\end{theorem}
For the maze network the analogous check covers 80 local configurations of a cell (wall, goal, and the reached bits of
the cell and its four neighbours). It gives output $\min(d(v),T)$ on every grid of every size. Its accumulators grow
by at most 6 per step (by 3 at the goal end of a corridor), and the readout's sum of absolute terms by at most 12 per
step, so float32 is exact while $12T\le2^{24}$. For the gene finder, a breadth-first search over (RNN state, codon)
pairs finds 3 reachable states and checks every transition, so it certifies every length (App.~\ref{app:float}).

\paragraph{Optimal-statistic tier.} Two short propositions (App.~\ref{app:props}) show that the written motion
network computes the Bayes rule and that the written Hamming decoder equals block ML except near ties.

\subsection{T3: no data}

Write the first layer as $W$, and call input coordinate $k$ \emph{unexcited} if $x_k=0$ on every training example.
Part (b) is standard non-identifiability.
\begin{theorem}[Unseen symbols and dead hinges]\label{thm:t3}
\begin{enumerate}
\item[(a)] Train, in exact arithmetic from zero-initialised optimiser states, with a data-independent learning-rate schedule and SGD, momentum SGD (with or without L2 decay), Adam without weight decay, or AdamW (decoupled
decay). Then the column $W_{:,k}$ of an unexcited coordinate ends at $c_TW_{0,:,k}$, where $c_T$ depends only on the
optimiser settings ($c_T=1$ without decay). So this column does not depend on the training data, and more data from
the same distribution does not move it. Excluded: Adam with coupled L2 decay (its per-entry scaling breaks $W_T=c_TW_0$, though the column still ignores the data) and global clipping with L2 inside the gradient.
\item[(b)] Let two targets $f,f'$ agree on $[0,a]$ and differ beyond a kink $K>a$ by a slope change $\Delta r$: $f'(x)-f(x)=\Delta r(x-K)$ for $x>K$. A
deterministic learner given examples with inputs in $[0,a]$ returns the same $\hat f$ for both; a randomised learner
returns the same distribution. So for every $x>K$, $\max\bigl(\mathbb E|\hat f(x)-f(x)|,\mathbb E|\hat f(x)-f'(x)|\bigr)
\ge\tfrac12|\Delta r|(x-K)$, and the error is unbounded for at least one of the two targets. For a one-hidden-layer
ReLU network trained by any optimiser in (a), a unit whose pre-activation is $\le0$ on every training input gets zero
gradient (in exact arithmetic, with the same decay on weights and bias, it stays dead while every $c_t>0$, e.g.\ SGD or AdamW with $\eta\lambda<1$; see the proof), so the hinges such units place beyond $a$ are set by initialisation, not by data.
\end{enumerate}
\end{theorem}
Part (b) holds for any learner; part (a) is about first-order training. For the day of week, the leading-digit
columns of 1600--1899 and 2100--2399 are unexcited when training on 1900--2099. For tax, the 2024 brackets above
\$500k gross income are never active in a \$0--500k training range. Roman numerals are \emph{not} an instance of (a):
M occurs in training, inside CM. What is missing is every window that starts with M or puts M before M, C or D. That
is an out-of-support failure, and we have no theorem for it.

\subsection{T4: no signal}

Let $d\in\Z_{10}^n$ be uniform and $\tau=(0,2,4,6,8,1,3,5,7,9)$ the Luhn ``double and add digits'' map. For a binary
word $b\in\{0,1\}^n$ let $h_b(d)=\mathbf 1\bigl[\sum_p\sigma_{b_p}(d_p)\equiv0 \pmod{10}\bigr]$ with
$\sigma_0=\mathrm{id}$ and $\sigma_1=\tau$. Luhn is $b=(1,0,1,0,\dots)$. Write $\tilde h=h-\tfrac1{10}$.
Part (a) mirrors a familiar parity fact; (b, c) are ours, (c) following \citet{shalevshwartz2017failures}.
\begin{theorem}[Luhn-type checksums carry no low-order signal]\label{thm:t4}
\leavevmode
\begin{enumerate}
\item[(a)] For every $A\subsetneq[n]$, $P(h_b=1\mid d_A)=\frac1{10}$: $h_b$ is independent of every proper subset of the digits.
\item[(b)] If $b,b'$ agree in at least one position and differ in a set $\Delta\ne\emptyset$, then
$\langle\tilde h_b,\tilde h_{b'}\rangle=\frac{5}{100}(0.2)^{|\Delta|}$ exactly, while $\|\tilde h_b\|^2=0.09$.
\item[(c)] Let $C\subseteq\{0,1\}^n$ be a code with a common fixed coordinate and minimum distance $\delta n$, and let
$p_w$ be differentiable with $\mathbb E\|\nabla_wp_w(d)\|^2\le G^2$. Then the gradients of $F_b(w)=\tfrac12\mathbb E(p_w(d)-h_b(d))^2$ satisfy
$\mathrm{Var}_{b\sim C}\nabla F_b(w)\le G^2\bigl(0.09/|C|+0.05\cdot0.2^{\delta n}\bigr)$. For any fixed
$0<\delta<\tfrac12$ a greedy Gilbert--Varshamov construction that starts from Luhn's $b$ gives such a code with
$|C|\ge2^{n-1}/\sum_{i<\delta n}\binom{n-1}{i}=2^{\Omega(n)}$. So if $G^2$ grows at most polynomially in $n$, the
gradient at any $w$ varies exponentially little across the family, so it says little about which scheme labels the data.
\end{enumerate}
\end{theorem}
\Cref{thm:t4} is about uniform digits and about learning which member of a family labels the data. The suite's GD
runs train one fixed target, Luhn, on a balanced mix of valid and invalid numbers. That GD stays at chance is
\emph{consistent with} the theorem; the theorem does not imply it. Two checks bring the setting closer. Under the
training mix, part (a) still holds (with the mix's class prior in place of $\frac1{10}$) for the two negative kinds that change one digit (80\% of negatives); only the
swap negatives carry a tiny pairwise signal (\cref{rem:mix}). And GD trained on the theorem's own distribution,
uniform digits with square loss, also stays at chance (\cref{sec:gd}). The written network holds every family member
with $S_{\max}+3$ hidden units of one shared row, so the obstacle is learnability, not expressivity.

\paragraph{What we do not prove.} Compounding error (recurrences run beyond the training size) has no theorem here. Out of support (Roman
$\ge$1000) has no theorem. The class we call data or recipe inefficiency has no theorem. We had a conjecture for Sudoku; tuned
GD is 100\% on every tested kind of Sudoku grid, so we withdrew it.


\section{Results}
\label{sec:results}

\subsection{Twelve networks are certified exact, one within a bound}
\label{sec:certs}

All rows of \cref{tab:suite} pass their certificates. Key numbers:
\begin{itemize}
\item \textbf{Enumeration rows.} ALU, day of week and routing are exact on every input; the tic-tac-toe policy loses no game. The largest
per-layer $M_i$ is 1,279 (ALU), $1.30\times10^6$ (day of week, grid $2^{-4}$), 5,061 and 131,072, all below
$2^{24}$. With the first-written constant 4, the day-of-week certificate returns 0 correct dates, all off by one day.
The tic-tac-toe policy ties in 33 of 694 positions; branching over every tied move, it still loses none of 1,264
games.
\item \textbf{Roman numerals.} The convolution and the sum over positions are exact by \cref{thm:t1}. For the
readout, the stored network's exact error plus the float32 bound of \cref{lem:margin} is at most 0.0088 in value
units on every numeral, against a rounding margin of 0.5 (57 times larger).
\item \textbf{DP, template, SMT and BFS rows.} Each certificate works in exact integers. Luhn's DP finds 145 states
and certifies all $10^{16}$ strings in 0.11\,s; its layers have $\max M_i=289$ and 44,405. Sudoku has $\max M_i\le$
2,435; the hydrogen unit 8; the chess network 78,085 (bounded over every 0/1 board, so even without the legality
constraints); the gene finder 10. z3 used the integer sort. So \cref{thm:t1} carries each certificate to float32.
\item \textbf{Tax (2024 rate schedule, single and joint, standard deduction).} In exact rationals the network is
within \$0.021 of the schedule on all of [\$0, \$2M]. Adding \cref{lem:margin}, the float32 network is within \$1.90
(single) and \$1.84 (joint) for every real income and every summation order. In PyTorch's order it is within \$0.18
at every whole dollar. The first-layer status gate (a bias of $-30$ cancelled by $+30$) costs most of this precision.
\item \textbf{Size-free networks.} The shortest-path update equals $\max(c,m)$ on all $1.7\times10^7$ integer pairs,
and $2B+L=8{,}197\le2^{24}$. The maze passes all 80 local configurations.
\end{itemize}

\subsection{Where the networks break}
\label{sec:float}

\begin{table}[t]
\caption{Failure points: logged prediction, its kind, and the measurement. \emph{Sharp}: derived from
\cref{thm:t2} or exact algebra. \emph{Band}: a heuristic range. Maze: a 40-cell corridor.}
\label{tab:x3}
\centering\footnotesize
\setlength{\tabcolsep}{2.5pt}
\resizebox{\columnwidth}{!}{%
\begin{tabular}{@{}llll@{}}
\toprule
network, setting & prediction & kind & measured \\
\midrule
shortest path, $\ell=5$, fp32 & first wrong hop 820 & sharp & 820 \\
shortest path, $\ell=1$, fp32 & right at 4096, wrong 4097 & sharp & 4097 \\
shortest path, fp16 / bf16 & wrong from hop 1 & sharp & 1 / 1 \\
maze, fp16 & $T^\star\in[512,700]$ & band & 684 \\
maze, bf16 & $T^\star\in[64,90]$ & band & 88 \\
maze, fp32 & none for $T\le3{,}000$ & sharp & none \\
tax gate, single & errors start at \$3,029,200 & sharp & \$3,029,200 \\
tax gate, joint & errors start at \$3,014,600 & sharp & \$3,014,600 \\
\bottomrule
\end{tabular}}
\end{table}

\Cref{tab:x3} and \cref{fig:x3} (App.~\ref{app:float}) give the predictions and the measurements. Six predictions
are sharp and all held.
In float16 the spacing above 2048 is 2, so $4096-5$ already rounds
at hop 1. The two maze predictions were only bands. Their lower ends assumed the goal accumulator grows by 4 per step;
on a corridor it grows by 3, and their upper ends were guesses. After the run, an exact-integer run of the corridor
(App.~\ref{app:float}) shows that the first stored value a 16-bit format cannot hold appears at $T=684$ (float16) and
$T=88$ (bfloat16), exactly where the errors start. This derivation came after the measurement.

\subsection{Hardened gradient descent}
\label{sec:gd}

\begin{table}[t]
\caption{Written network against tuned GD on the same architecture (selected on validation, then 3 seeds). Ranges are
over seeds; ``train'' gives training sizes; test sizes and intervals are in App.~\ref{app:gd}. Chance: 50\% (Luhn), 0.5 AUC, 14.3\% (day of week).}
\label{tab:gd}
\centering\footnotesize
\setlength{\tabcolsep}{2pt}
\resizebox{\columnwidth}{!}{%
\begin{tabular}{@{}lllc@{}}
\toprule
task, test & written & tuned GD & pred.\ held \\
\midrule
Luhn, mix of 4 kinds (train 10.2M) & 100 & 49.3--51.0 & yes \\
Luhn, uniform digits, AUC (train 10.2M) & 1.0 & 0.499--0.503 & yes \\
Roman, random split & 100 & 95.8--97.3 & yes \\
Roman, train 1--999, test $\ge$1000 & 100 & 0.0 & yes \\
day of week, unseen centuries, one-hot & 100 & 6.7--15.4 & yes \\
day of week, unseen centuries, scalar & 100 & 0.6--2.9 & yes \\
ALU, 1\% data: ADD / SUB & 100 & 42.2--54.5 / 44.3--51.5 & yes \\
maze $30\times30$ ($\times4$; 98.6--99.7 at $10\times10$) & 100 & 0.005--0.15 & yes \\
maze $10\times10$ ($\times1$) & 100 & 25.0--51.6 & no \\
Sudoku, every kind ($\times4$ and $\times1$) & 100 & 100.0 & \textbf{no} \\
routing (train 3k packets) & 100 & 98.2--98.5 & no \\
chess (train 20k positions) & 100 & 88.9--89.1 & no \\
Hamming, BER at 4\,dB (train 10k words) & 0.531 & 0.619--0.646 & in part \\
chemistry: H counts, 100 atoms & 100 & 100 & yes \\
chemistry: mass error, 100 atoms (Da) & 0.0001 & 0.17--0.31 & yes \\
\bottomrule
\end{tabular}}
\end{table}

The comparison is not symmetric: the writer reads the specification, GD sees only data. It asks what learning
from data misses when a specification exists, not whether writing can replace GD without one. \Cref{tab:gd}
summarises the tuned baselines.
\begin{itemize}
\item \textbf{No signal (Luhn).} GD stays at chance in every configuration and on every test kind, including
curriculum and $4\times$ width, with 10.24M fresh numbers per seed. With uniform digits and square loss, the setting
of \cref{thm:t4}, the test AUC is 0.499--0.503 on 200,000 strings. Both are consistent with \cref{thm:t4}.
\item \textbf{No data.} Trained on 1900--2099, day of week gets 6.7--15.4\% on unseen centuries with one-hot century
digits. One could blame the one-hot input, whose unseen columns never move (\cref{thm:t3}(a)). With a single scalar
century input, which \emph{is} active in training, GD does no better: 0.6--2.9. This matches the
non-identifiability behind \cref{thm:t3}(b), in a classification analogue: two centuries of data do not fix the others. Trained on Roman numerals 1--999, every seed gets 0\%
of 1000--3999, an out-of-support failure without a theorem. Untuned GD tax networks under-predict above their
\$500k training range (App.~\ref{app:gd}); \cref{thm:t3}(b) says only that at least one of two targets that agree up to \$500k is extrapolated wrongly, not the sign.
\item \textbf{Compounding error.} The $4\times$-wide maze network with a curriculum is nearly exact at training size
but gets 0.005--0.15\% of cells right at $30\times30$. At the written width ($\times1$) GD already fails at training
size (25--52\%), so there the failure is not compounding error; we assign this class at $\times4$ only.
\item \textbf{Data or recipe inefficiency.} Sudoku is the key negative. With learning rate $3\times10^{-4}$, AdamW and
30,000 steps of fresh grids, every seed is 100\% on every kind at both widths, so we withdrew an earlier claim that
GD cannot find this rule. Tuned GD with the notebook's middle data budgets stays below the written networks on
routing (98.2--98.5\%), chess (88.9--89.1; random boards, 47\% checks), Hamming (BER 0.62--0.65\% against 0.53\% at 4\,dB) and the chemistry
masses (0.17--0.31\,Da against $10^{-4}$), while it learns the hydrogen rule exactly. With ten to a hundred times
more data the notebook's untuned GD reached 100\% on routing; we did not tune that setting.
\end{itemize}

\begin{table}[t]
\caption{Evidence per task. Tuned: a GD baseline tuned in this paper. Theorem: a theorem in this paper whose setting
covers the task.}
\label{tab:evidence}
\centering\footnotesize
\setlength{\tabcolsep}{2pt}
\resizebox{\columnwidth}{!}{%
\begin{tabular}{@{}lllll@{}}
\toprule
class & task & tuned & theorem & both? \\
\midrule
no signal & Luhn & \checkmark & T4 (family; consistent) & yes, in part \\
no data & day of week & \checkmark & T3(a) one-hot, T3(b)-type scalar & yes \\
no data & tax above \$500k & -- & T3(b) & theorem only \\
out of support & Roman $\ge$1000 & \checkmark & -- & tuned only \\
compounding & maze $30\times30$ ($\times4$) & \checkmark & -- & tuned only \\
compounding & shortest paths, gene finder & -- & -- & neither \\
inefficiency & ALU, routing, chess, Hamming, chem. & \checkmark & -- & tuned only \\
(solved) & Sudoku & \checkmark & -- & GD succeeds \\
\bottomrule
\end{tabular}}
\end{table}

\Cref{tab:evidence} applies our own rule: a GD failure is a mechanism only when a tuned baseline and a theorem both
support it. Only no signal (in part) and no data have this partial theoretical support. Compounding error, out-of-support and inefficiency are labels for what we
observed, not shown mechanisms.

\subsection{Fresh specifications}
\label{sec:x6}

\begin{table}[t]
\caption{Thirty fresh specifications, first submissions. Certified: exact for any summation order. Tax: the pre-fixed
rule is at most \$1 error at every whole dollar \$0--2M, both statuses, float32. 95\% LB: one-sided Clopper--Pearson.}
\label{tab:x6}
\centering\footnotesize
\setlength{\tabcolsep}{2pt}
\resizebox{\columnwidth}{!}{%
\begin{tabular}{@{}llll@{}}
\toprule
family & result & certificate $\to$ float32 & 95\% LB \\
\midrule
check digits ($m$ 7--13, $n$ 8--14) & 10/10 certified & DP over $10^8$--$10^{14}$, T1 & 74\% \\
routing (16--45 routes) & 10/10 certified & enum over $2^{18}$--$2^{20}$, T1 & 74\% \\
calendars (week 5--10) & 5/5 certified & all $\approx$292k dates, T1 & 55\% \\
tax (4--8 brackets) & 5/5 pass the rule & every \$1 in one run & 55\% \\
 & \quad 0/5 exact to the cent & max error \$0.07--0.11 & \\
 & \quad 1/5 certified within \$1 & L1 bound \$0.94--2.63 & \\
\midrule
total & 25/30 certified; 30/30 pass & & 68\%; 90.5\% \\
\bottomrule
\end{tabular}}
\end{table}

Twenty-five first submissions are certified exact for any summation order (\cref{tab:x6}). All five tax submissions
pass the rule fixed in the scorer before the sessions: every whole dollar from \$0 to \$2M, both statuses, float32,
error at most \$1. Their measured errors are \$0.07--0.11. Under a strict rule, exact to the cent, all five fail, and
the count is 25/30. In that float32 run on whole dollars, any tolerance of at least \$0.12 gives 30/30. \Cref{lem:margin} certifies only one tax network
within \$1 for every order; the rest are tested, not certified, at that tolerance. The bounds (68\% for 25/30, 90.5\%
for 30/30) treat the 30 specifications as independent, but they come from four families; the
per-family bounds in \cref{tab:x6} are the safer reading.

We predicted at least 80\% and failures in derived constants. Accuracy exceeded 80\%, and the predicted failure mode did not appear. The transcripts show one
reason: the writers derived every offset, kink and table from the spec in Python. Constructions repeat across
sessions. All ten check-digit writers reduce each digit's contribution modulo $m$ and read the sum with a ``comb'' of
one-integer bumps. Routing writers use output weights $2^{\text{prefix length}}$ (six) or a parent-cancellation
scheme (four). Tax writers solve for the kink where tax meets the credit. Calendar writers split the day count into
per-field tables modulo the week length: new instances of familiar families (limitation (ii)).

\subsection{Second writer family}
\label{sec:x6b}

\begin{table}[t]
\caption{Fresh specifications by two writers, rules of \cref{tab:x6}.}
\label{tab:x6b}
\centering\footnotesize
\setlength{\tabcolsep}{3pt}
\begin{tabular}{@{}lccccc@{}}
\toprule
writer & routing & check digit & tax & calendar & total \\
\midrule
Qwen, basic & 10/10 & 9/10 & 4/5 & 0/5 & 23/30 \\
Qwen, expanded & 10/10 & 9/10 & 5/5 & 1/5 & 25/30 \\
Claude & 10/10 & 10/10 & 5/5 & 5/5 & 30/30 \\
\bottomrule
\end{tabular}
\end{table}

Qwen3.8-27B (local Ollama, unsloth GGUF revision \texttt{4af51b35}) wrote the same 30 specifications twice. The
basic run (UD-Q4\_K\_S) had one generation, a syntax-only retry, no code execution and a 14k thinking cap, hit in
every session. It passes 23 (95\% CI 0.58--0.90; paired exact McNemar $p=0.016$ against Claude; predicted 10--22).
The expanded run (UD-Q6\_K) allowed 40k thinking tokens and ran each script, returning Python errors (never scorer
output) for up to two corrections. It passes 25 (CI 0.65--0.94; $p=0.063$; predicted 24--28); all five calendar
sessions hit the cap. This narrows the gap by two but does not separate quantisation, thinking and run feedback.

\subsection{Bayes gaps and mutation testing}
\label{sec:bayes}

The written Hamming decoder matches block ML to within 0.07\% relative at every SNR and differs from it on at most
0.011\% of words; bitwise MAP has 1.8\% fewer bit errors at 0\,dB (\cref{tab:x7}, App.~\ref{app:checks}). The motion
detector is within one standard error of the exact Bayes accuracy everywhere and 9.9 points above GD trained on
60,000 pairs at coherence 0.2. In mutation testing (App.~\ref{app:checks}) the certificates decided every
single-weight mutant in exact arithmetic, and random tests of 20,000 inputs missed 42 of 567 (7.4\%) of the mutants that break exactness.


\section{Discussion and limitations}
\label{sec:discussion}

\paragraph{Why certificates are cheap here.} Written networks use small integer or binary-fraction weights and few
active units, so their reachable state sets are tiny: 145 Luhn states, 3 gene-finder states. ReLU verification
is hard in general \citep{katz2017reluplex,wang2021beta}, even with weights in $\{-1,0,1\}$; ours is cheap because
short binary weights make float32 transfer trivial and the author's structure keeps the check small. Certificates also say where a network breaks, which tests do not: the shortest-path network passed every test up
to 200 nodes and is wrong from distance 4,097.
\paragraph{Limitations.} (i) The suite comes from one model family; an open model passes 25 of 30 fresh specifications
(\cref{sec:x6b}).
(ii) The new specifications are new instances of known families: they rule out recall of specific circuits, not
familiarity with the idioms. The suite has one network per task, so we measure first-try success, not variance
across sessions. (iii) Each writer saw the harness context, a 14-line memory index about the
programme and a prompt that hints at per-digit tables. Writers author Python; we ran no ablation without code.
(iv) \Cref{thm:t4} is family-level and for uniform digits. (v) We certify no transformer, and no network whose task
lacks an exact or closed-form Bayes specification; the shortest-path and gene-finder baselines were not tuned. (vi)
The prediction log is self-attested.

\paragraph{Compute, LLM use and release.} All runs used CPU (about 9 core-hours) except the two Qwen runs (4.4 and
4.8 GPU-hours; 513k and 1.05M generated tokens). The 30 Claude sessions used about 1.37M tokens. The text was drafted with LLM assistance and checked against the logged results.
We will release all weights, code, prompts, transcripts and results; re-checking needs only a CPU.


\phantomsection\label{bodyend}
\section*{Impact statement}
This work makes small neural networks written by language models checkable: a certificate says on which inputs, and
in which arithmetic, the network is exact, and where it stops being exact. That can make such networks safer to use
inside larger systems. The risks are the usual ones for automation of rules. A certified network is only as correct
as its specification; a wrong specification (for example an outdated tax schedule) is certified just as readily.
Outside its stated domain a network fails silently, so the operating envelope and its guard must ship with it. We
see no direct path to misuse beyond that of the rules being encoded.


\setlength{\bibsep}{2pt plus 1pt}

\newpage
\appendix
\onecolumn
\section{Proofs}
\label{app:proofs}

We give full proofs, including those of standard facts, so that the proofs behind the certificates can be checked from this paper
alone. Where a step is known, we cite its source in \cref{sec:theory}.

\subsection{Proof of \cref{thm:t1}}
Binary32 has a 24-bit significand, so every integer $N$ with $|N|\le2^{24}$ is representable, and so is $N2^{-e}$ for
any $e\in[-103,149]$ (using subnormals for $e>126$; all our grids have $e\le45$). Fix layer $\ell$, unit $i$ and an input in
$\mathcal D$. Each product $W_{ij}x_j=(\hat W_{ij}\hat x_j)2^{-e_\ell}$ lies on the grid $2^{-e_\ell}\Z$ and has
$|\hat W_{ij}\hat x_j|\le M_i\le2^{24}$, so it is representable and the rounded product equals it. Any partial sum of
a subset of the products, plus possibly the bias, is a grid point of integer magnitude at most $M_i\le2^{24}$. So it is
representable, and round-to-nearest returns it unchanged. A fused multiply--add rounds the exact value once, so the
same holds. Hence every intermediate value is exact, in every order. ReLU and max are exact; a sum over positions is a
layer with unit weights, covered by the same argument. The outputs of layer $\ell$ lie on $2^{-e_\ell}\Z$, the input
grid of layer $\ell+1$, and induction over layers finishes the proof. If any operation is done in a narrower format
(TF32, float16, bfloat16) or by an algorithm that forms other intermediate values (Winograd or FFT convolution), the
argument does not apply. \qed

\subsection{Proof of \cref{lem:margin}}
Fix layer $\ell$ and assume the claim for $\ell-1$. The float32 layer computes $\hat z_\ell=\mathrm{fl}(W_\ell\hat
h_{\ell-1}+b_\ell)$, an inner product of $n_\ell$ terms (the bias times 1 included). For any summation order, with or
without fused multiply--add, the standard bound \citep[Sec.~3.1]{higham2002accuracy} gives
$|\mathrm{fl}(W\hat h+b)-(W\hat h+b)|\le\gamma_{n}(|W||\hat h|+|b|)$ entrywise. Hence
\[|\hat z_\ell-z_\ell|\le\gamma_{n_\ell}(|W_\ell||\hat h_{\ell-1}|+|b_\ell|)+|W_\ell||\hat h_{\ell-1}-h_{\ell-1}|.\]
By induction $|\hat h_{\ell-1}-h_{\ell-1}|\le\epsilon_{\ell-1}$, so $|\hat h_{\ell-1}|\le|h_{\ell-1}|+\epsilon_{\ell-1}$,
which gives $|\hat z_\ell-z_\ell|\le\epsilon_\ell$. ReLU is 1-Lipschitz and exact in float32, so $|\hat h_\ell-h_\ell|\le
\epsilon_\ell$. If $z+\epsilon\le0$ then $\hat z\le0$ and $\hat h=0$ exactly; adding an exact zero is exact, so such
a unit is not a rounded term of the next layer. Finally, if the exact output $o$ satisfies $o-\theta>\epsilon_L$ (or
$\theta-o>\epsilon_L$), then $\hat o$ lies on the same side of the threshold $\theta$. \qed

\emph{Which weights.} In the lemma and in every application, $W_\ell$ and $b_\ell$ are the \emph{stored float32}
weights read as exact rationals (\texttt{Fraction(float(w))} in \texttt{code/x9\_float32.py}), not the decimal design
values (e.g.\ $0.01$). So the ``exact'' network is the stored network evaluated without rounding, and no
weight-rounding term is needed: any gap between the stored weights and the design values is already inside the exact
error that we compare with the specification below.

\emph{Applications.} For the Roman reader, the convolution and the sum over positions are exact by \cref{thm:t1}
($\max M_i=3$ and 4), so only the readout (33 terms) rounds: $\epsilon=\gamma_{33}(|w|^\top h+|b|)$, evaluated exactly for
each of the 3,999 numerals. For the tax networks the input $x=I/10^5$ is itself rounded once to float32 (relative error
at most $2^{-24}+2^{-52}$). We use $|\hat f(\hat x)-f(x)|\le|\hat f(\hat x)-f(\hat x)|+|f(\hat x)-f(x)|$. The second term is at most
$\mathrm{Lip}\cdot|x-\hat x|$ with $\mathrm{Lip}=\sum_j|w_{2j}w_{1j0}|$; we bound the first, including the choice of
units that cannot switch on, as a maximum over all $\hat x\in[0,x_{\max}(1+2^{-24}+2^{-52})]$, so it holds at the
rounded input. We also add the float64 rescaling by $10^5$. Underflow can add less than $10^{-35}$ dollars, below
the rounding of the reported bounds. Both $\epsilon$ and the exact error are piecewise linear in $x$, so their maxima over
an interval are attained at hinges, at points where a unit can switch on in float32, and at the ends. We evaluate
them there in exact rational arithmetic (App.~\ref{app:float}).

\subsection{Proof of \cref{thm:t2}}
Let $d_t(v)$ be the length of a shortest walk from $s$ to $v$ with at most $t$ edges ($\infty$ if none). So $d_0=0$
at $s$ and $\infty$ elsewhere, and $d_{t+1}(v)=\min\bigl(d_t(v),\min_{u\in N(v)}d_t(u)+w_{uv}\bigr)$. We show
$c_t(v)=\max(0,B-d_t(v))$ by induction. The base case is the initialisation. Suppose it holds at $t$. Then
$c_t(\cdot)\in[0,B]$, and since $w\in[1,L]$, every message $c_t(u)-w_{uv}$ lies in $[-L,B]$. So the local check applies
and $c_{t+1}(v)=\max(c_t(v),m_t(v))$; for an isolated node $m_t=0$ and $c_{t+1}=c_t$. Expanding,
\[c_{t+1}(v)=\max\Bigl(0,\;B-d_t(v),\;\max_u\bigl(-w_{uv}\bigr),\;\max_u\bigl(B-d_t(u)-w_{uv}\bigr)\Bigr).\]
The third group is negative, so the leading $0$ dominates it. The rest is $\max(0,B-d_{t+1}(v))$. The output is
$B-c_T(v)=\min(d_T(v),B)$, and $d_T=d$ once $T\ge|V|-1$. On a path with edge length $\ell$, hop $k$ has $d=k\ell$,
which is exact iff $k\ell\le B$, so the first wrong hop is $\lfloor B/\ell\rfloor+1$.

\emph{Float32.} The update unit computes $\mathrm{relu}(-(p+q)+m)$ with $p,q\ge0$ and $p+q=c\le B$, so its sum of
absolute terms is at most $p+q+|m|\le B+\max(B,L)\le2B+L$. The message $c+w_{uv}\cdot(-1)$ has at most $B+L$, and the readout
$B-(p+q)$ at most $2B$. All weights are integers, so \cref{thm:t1} applies whenever $2B+L\le2^{24}$ (here
$8{,}197$).

\emph{Reduction of the shortest-path network to \eqref{eq:upd}.} The structure check verifies: unit 0 receives
$\mathrm{relu}(p+q)$ and unit 1 receives $\mathrm{relu}(-(p+q)+m_1)$; all other units are identically zero; the
message of unit 1 is $p+q$ with edge coefficient $-1$; the message of unit 0 is $0$; and the readout is $B-(p+q)$ with
$p_0=B\mathbf 1[v=s]$, $q_0=0$. So $c=p+q$ follows \eqref{eq:upd} with $F(c,m)=\mathrm{relu}(c)+\mathrm{relu}(m-c)$,
and the finite check confirms $F=\max$ on the grid.

\emph{Maze.} Let $r_t(v)=u_0-u_1\in\{0,1\}$. Both channels share their input and recurrent weights and differ in bias
by 1, so $r_t(v)=\mathrm{clamp}(z,0,1)$ for the shared pre-activation $z$. This depends on the five-cell
neighbourhood only through the $r$ values of those cells, $\mathrm{goal}(v)$ and $\mathrm{wall}(v)$. The
80-configuration check verifies $r_{t+1}(v)=\mathrm{free}(v)\wedge(\mathrm{goal}(v)\vee r_t(v)\vee\bigvee_{u\sim
v}r_t(u))$, and walls are never reached. By induction $r_t(v)=\mathbf 1[\mathrm{free}(v)\wedge d(v)\le t-1]$, and the
readout $\sum_{t=1}^T(\mathrm{free}(v)-r_t(v))=\min(d(v),T)$ on reachable free cells. Since $u_0=\mathrm{relu}(\mathrm{goal}+\sum
r)\le6$ and $u_1\le5$, the accumulators grow by at most 6 and 5 per step and the free channel by 1. The readout's sum
of absolute terms grows by at most 12 per step, so \cref{thm:t1} holds while $12T\le2^{24}$, i.e.\ $T\le1{,}398{,}101$.
\qed

\subsection{Statements of the two propositions}\label{app:props}
\begin{proposition}[Motion correlator is Bayes]\label{prop:motion}
In the motion generator (frame 1 i.i.d.\ Bernoulli$(\rho)$ on a torus; frame 2 its shift by $d$, each pixel replaced
with probability $1-\kappa$ by a fresh Bernoulli$(\rho)$ dot; uniform prior on 9 shifts), the log-likelihood of shift
$d$ is $n_{11}(d)\log\frac{P_{11}P_{00}}{P_{10}P_{01}}+\mathrm{const}$, where $n_{11}(d)$ counts pixels on in frame 2
whose $d$-preimage is on in frame 1. For $\kappa,\rho\in(0,1)$ the log-odds factor is positive, so the Bayes rule is
$\arg\max_dn_{11}(d)$, which the written network computes. At $\kappa=1$ its maximisers are the shifts consistent with the frames, so it is again Bayes.
\end{proposition}
\begin{proposition}[What the written Hamming decoder is]\label{prop:ham}
For BPSK in AWGN \citep{lin2004error} with uniformly distributed codewords, $p(c\mid y)\propto\exp(\langle y,s_c\rangle/\sigma^2)$, where $\sigma^2=1/(2R\,E_b/N_0)$ with code rate
$R=4/7$. The written decoder takes a softmax of $\beta\langle y,s_c\rangle$ with $\beta=30$ and thresholds each bit's
posterior marginal. Its bit marginals are the exact posterior marginals for all $y$ iff $\beta=1/\sigma^2$ ($2.87$ at $E_b/N_0=4$\,dB), and it tends to block ML
(nearest codeword, unique almost surely) as $\beta\to\infty$. It agrees with block ML whenever the top-two correlation gap exceeds
$\ln15/\beta\approx0.0903$.
\end{proposition}

\subsection{Proof of \cref{prop:motion}}
Given frame 1 and shift $d$, the pixels of frame 2 are independent with $P(f_2[x]=1)=\kappa+(1-\kappa)\rho$ if
$f_1[x-d]=1$ and $(1-\kappa)\rho$ otherwise. With $n_{ab}(d)=\#\{x:f_1[x-d]=a,f_2[x]=b\}$,
$\log p(f_2\mid f_1,d)=\sum_{a,b}n_{ab}(d)\log P_{ab}$. On the torus $n_{10}+n_{11}=|f_1|$ and $n_{01}+n_{11}=|f_2|$ do
not depend on $d$, and $n_{00}=N^2-|f_1|-|f_2|+n_{11}$. Substituting leaves
$n_{11}(d)\log\frac{P_{11}P_{00}}{P_{10}P_{01}}$ plus terms constant in $d$. The odds ratio exceeds 1 because
$P_{11}>P_{01}$ whenever $\kappa>0$. With a uniform prior the Bayes rule is $\arg\max_dn_{11}(d)$. Each written unit
is $\mathrm{relu}(f_1[x]+f_2[x+d]-1)=f_1[x]f_2[x+d]$ on binary frames, so its image mean is $n_{11}(d)/N^2$. \qed

\subsection{Proof of \cref{prop:ham}}
With $y=s_c+\sigma\xi$, $\log p(y\mid c)=-\|y-s_c\|^2/(2\sigma^2)+\mathrm{const}=\langle y,s_c\rangle/\sigma^2+\mathrm{const}$,
since $\|s_c\|^2=7$ for all $c$. Energy per information bit $E_b=7/4$ times the energy per symbol, so
$\sigma^2=1/(2RE_b/N_0)$ with $R=4/7$. With a uniform prior the posterior is $\mathrm{softmax}(\langle y,s_c\rangle/\sigma^2)$,
and bitwise MAP thresholds $\sum_cp(c\mid y)m_{c,i}$ at $\tfrac12$. The written network computes the same with
$1/\sigma^2$ replaced by $\beta$. As $\beta\to\infty$ the softmax concentrates on the nearest codeword. If the top
correlation exceeds the second by $g$, the top codeword gets weight at least $1/(1+15e^{-\beta g})$, which is above
$\tfrac12$ when $g>\ln15/\beta$; then every bit marginal agrees with that codeword's bit. \qed

\subsection{Proof of \cref{thm:t3}}
(a) On a training input $x_k=0$, so the gradient of the loss with respect to the column $W_{:,k}$ (for a convolution,
the kernel slice reading channel $k$) is $\delta\,x_k^\top=0$ for every example and every batch. Plain SGD makes no
update. With L2 decay folded into the gradient, SGD and momentum SGD update the column by a linear recurrence with
fixed coefficients, so it ends at $c_TW_{0,:,k}$. Adam's moment estimates stay at zero and its update
$\hat m/(\sqrt{\hat v}+\epsilon)$ is zero. AdamW's decoupled decay multiplies by $(1-\eta\lambda)$ per step. In all
cases $c_T$ depends only on $(\eta,\mu,\lambda,T)$.
With coupled L2 decay, Adam's gradient for the column is $\lambda W_{:,k}\ne0$, and the normalised update moves each
entry by about $\eta$ per step regardless of its size; so that optimiser is excluded. The other parameters' forward
values on training inputs do not involve $W_{:,k}$, so their trajectory depends only on their initialisation, the data
and the batch order. The column itself never depends on the data.

(b) Two training sets drawn from $[0,a]$ under $f$ and $f'$ have the same distribution as labelled samples. A
deterministic learner returns the same $\hat f$; a randomised one returns the same distribution over $\hat f$. For
each $x>K$, $|\hat f(x)-f(x)|+|\hat f(x)-f'(x)|\ge|f(x)-f'(x)|=|\Delta r|(x-K)$. Taking expectations, the larger of the
two expected errors at $x$ is at least $\tfrac12|\Delta r|(x-K)$. The bound may switch between $f$ and $f'$ as $x$
varies, but $\sup_x\max(\cdot,\cdot)=\infty$, so at least one of the two targets has unbounded error. For a dead unit,
$\mathrm{relu}'(z)=0$ at $z\le0$ (the PyTorch convention), so the gradients of its input weights and bias vanish, and the
gradient of its output weight is proportional to $\mathrm{relu}(z)=0$. Under decay that scales its input weights and
bias by the same $c_t$, its pre-activation on every training input is $c_t$ times the initial one, so it stays
$\le0$ while $c_t>0$. This holds for SGD and AdamW with $\eta\lambda<1$, where $c_t=(1-\eta\lambda)^t$. Momentum
SGD with coupled decay gives $c_t=(1+\mu-\eta\lambda)c_{t-1}-\mu c_{t-2}$, which oscillates in sign when
$\eta\lambda>(1-\sqrt\mu)^2$ ($0.0026$ at $\mu=0.9$); then a dead unit can switch on. Our GD runs use Adam or AdamW. \qed

\subsection{Proof of \cref{thm:t4}}
(a) Take $p\notin A$ and condition on all digits except $d_p$, which determines $d_A$. Then $S=\sigma_{b_p}(d_p)+R$
with $R$ fixed, and $\sigma_{b_p}$ is a bijection of $\Z_{10}$, so $S\bmod10$ is uniform and $P(h_b=1\mid
d_{-p})=\frac1{10}$. Averaging over the remaining digits gives the claim.

(b) Let $\omega=e^{2\pi i/10}$. From $\mathbf 1[S\equiv0]=\frac1{10}\sum_{j=0}^{9}\omega^{jS}$,
$\tilde h_b(d)=\frac1{10}\sum_{j=1}^{9}\prod_p\omega^{j\sigma_{b_p}(d_p)}$. The digits are independent, so
\[\langle\tilde h_b,\tilde h_{b'}\rangle=\frac1{100}\sum_{j,j'=1}^{9}\prod_p\mathbb E_x\,\omega^{j\sigma_{b_p}(x)-j'\sigma_{b'_p}(x)}.\]
At a position where $b_p=b'_p$ the factor is $\mathbb E\,\omega^{(j-j')\sigma(x)}=\mathbf 1[j=j']$, because $\sigma$ is a
bijection; so only $j=j'$ survives. At $p\in\Delta$ the factor is $\mathbb E\,\omega^{\pm jv(x)}$ with
$v(x)=x-\tau(x)\bmod10$. Listing $v$ over $x=0,\dots,9$ gives $0,9,8,7,6,4,3,2,1,0$: the value $0$ twice, each of
$1,2,3,4,6,7,8,9$ once, and never $5$. Hence $\sum_x\omega^{jv(x)}=\sum_{y\in\Z_{10}}\omega^{jy}-\omega^{5j}+1=1-(-1)^j$,
and the factor is $0.2$ for odd $j$ and $0$ for even $j$, for either sign. Summing the five odd $j$ gives
$\frac5{100}(0.2)^{|\Delta|}$. Finally $\|\tilde h_b\|^2=\mathrm{Var}(h_b)=0.1\cdot0.9$.

(c) $\nabla F_b(w)=\mathbb E[(p_w-\tfrac1{10})\nabla p_w]-\mathbb E[\tilde h_b\nabla p_w]$, and the first term does not
depend on $b$. Write $g=\nabla p_w$ with components $g_i$, and let $H$ be the Gram matrix of $\{\tilde h_b\}_{b\in C}$. Then
\[\mathrm{Var}_b\nabla F_b\le\frac1{|C|}\sum_b\bigl\|\mathbb E[\tilde h_bg]\bigr\|^2=\frac1{|C|}\sum_i\sum_b\langle\tilde h_b,g_i\rangle^2
\le\frac{\lambda_{\max}(H)}{|C|}\sum_i\|g_i\|^2\le\frac{\lambda_{\max}(H)}{|C|}G^2.\]
All codewords share the fixed coordinate, so (b) applies to every pair, and by Gershgorin
$\lambda_{\max}(H)\le0.09+(|C|-1)\,0.05\cdot0.2^{\delta n}$. For the code: start with Luhn's $b$ and add, greedily, any
word with the same first coordinate at distance at least $\delta n$ from all chosen words. While fewer than
$2^{n-1}/\sum_{i<\delta n}\binom{n-1}{i}$ words are chosen, the balls of radius $\delta n-1$ around them do not cover
the $2^{n-1}$ words with that first coordinate, so the greedy step succeeds; this is the Gilbert--Varshamov argument.
The bound is $2^{\Omega(n)}$ for fixed $\delta<\tfrac12$. With $G^2=\mathrm{poly}(n)$ the variance is exponentially small. \qed

\begin{remark}[\cref{thm:t4}(a) under the suite's training mix]\label{rem:mix}
The Luhn GD runs train on valid numbers and three kinds of invalid ones: a wrong check digit (40\%), one changed digit
(40\%) and a swap of two neighbouring digits (20\%). A valid number has any 15 of its 16 digits i.i.d.\ uniform.
Changing one digit $p$ to $d_p+r$ with $r$ uniform on $\{1,\dots,9\}$ and independent of the rest keeps any 15 digits
i.i.d.\ uniform. So for the first two kinds, every proper subset of digits has the same distribution in both classes,
and part (a) holds for a mix of valid numbers with these kinds, with the mix's class prior in place of $\frac1{10}$. A swap is kept only when it breaks validity, which
excludes equal neighbours at the swapped spot. So swap negatives carry a small pairwise signal: at a given neighbouring
pair, $P(d_p=d_{p+1})$ is about 0.1 percentage points lower among negatives than among valid numbers.
\end{remark}


\section{Float32 coverage of every certificate}
\label{app:float}

Every certificate reasons in exact integers or rationals. This appendix shows, row by row, why the conclusion also
holds for the float32 network as run, in any summation order. Code: \texttt{code/x9\_float32.py}; results:
\texttt{results/x9\_float32.json}.

\paragraph{The quantity checked.} For \cref{thm:t1} we compute, for every layer, $\max_iM_i$ with
$M_i=\sum_j|\hat W_{ij}\hat x_j|+|\hat b_i|$, the sum of absolute scaled terms in units of that layer's output grid
$2^{-e}$. It is the maximum over the whole domain (enumeration rows), over all reachable states (DP and BFS rows), or
a worst case over every input the encoding allows (template and SMT rows). It is not the largest pre-activation; the
earlier text was ambiguous about this, but the code always computed $M_i$. \Cref{tab:layers} lists every value. All
are at most $2^{24}=16{,}777{,}216$.

\begin{table}[h]
\caption{Per-layer $\max_iM_i$ and grid exponent $e$ for every certified network (suite and X6).}
\label{tab:layers}
\centering\scriptsize
\begin{tabular}{@{}lll@{}}
\toprule
network & certificate & per layer: $\max_iM_i$ (grid exponent $e$) \\
\midrule
8-bit ALU & enum & 766 ($e{=}0$); 1,279 ($e{=}0$); 30 ($e{=}0$) \\
day of week & enum & 663 ($e{=}2$); 5,352 ($e{=}4$); 1,297,920 ($e{=}4$) \\
tic-tac-toe & enum & 9 ($e{=}0$); 5,061 ($e{=}0$) \\
routing & enum & 27 ($e{=}0$); 131,072 ($e{=}0$) \\
Luhn & DP & 289 ($e{=}0$); 44,405 ($e{=}0$) \\
Sudoku validity & tmpl & 10 ($e{=}0$); 2,435 ($e{=}0$) \\
chemistry H count & tmpl & 4 ($e{=}0$); 8 ($e{=}0$); 4 ($e{=}0$) \\
chess check & SMT & 76 ($e{=}0$); 128 ($e{=}0$); 78,085 ($e{=}0$) \\
gene finder & BFS & 5 ($e{=}0$); 6 ($e{=}0$); 5 ($e{=}0$); 10 ($e{=}0$) \\
Roman (conv, pool) & enum & 3 ($e{=}0$); 4 ($e{=}0$) \\
shortest paths & local & 8,197 ($e{=}0$) \\
maze (bound per step $T$) & local & $12T$ ($e{=}0$) \\
X6 cd01 & DP & 179 ($e{=}0$); 2,688 ($e{=}1$) \\
X6 cd02 & DP & 120 ($e{=}0$); 2,241 ($e{=}0$) \\
X6 cd03 & DP & 128 ($e{=}0$); 2,081 ($e{=}0$) \\
X6 cd04 & DP & 208 ($e{=}0$); 4,680 ($e{=}1$) \\
X6 cd05 & DP & 161 ($e{=}0$); 2,627 ($e{=}0$) \\
X6 cd06 & DP & 195 ($e{=}0$); 5,200 ($e{=}1$) \\
X6 cd07 & DP & 252 ($e{=}0$); 6,672 ($e{=}1$) \\
X6 cd08 & DP & 155 ($e{=}0$); 3,479 ($e{=}0$) \\
X6 cd09 & DP & 113 ($e{=}0$); 1,794 ($e{=}1$) \\
X6 cd10 & DP & 228 ($e{=}0$); 5,952 ($e{=}1$) \\
X6 rt01 & enum & 32 ($e{=}0$); 2,105,344 ($e{=}1$) \\
X6 rt02 & enum & 28 ($e{=}0$); 4 ($e{=}0$) \\
X6 rt03 & enum & 27 ($e{=}0$); 2,097,152 ($e{=}1$) \\
X6 rt04 & enum & 27 ($e{=}0$); 1,048,577 ($e{=}1$) \\
X6 rt05 & enum & 29 ($e{=}0$); 532,480 ($e{=}1$) \\
X6 rt06 & enum & 29 ($e{=}0$); 4 ($e{=}0$) \\
X6 rt07 & enum & 27 ($e{=}0$); 5 ($e{=}0$) \\
X6 rt08 & enum & 27 ($e{=}0$); 4 ($e{=}0$) \\
X6 rt09 & enum & 32 ($e{=}0$); 2,097,153 ($e{=}1$) \\
X6 rt10 & enum & 31 ($e{=}0$); 2,099,200 ($e{=}1$) \\
X6 ca01 & enum & 23 ($e{=}0$); 50 ($e{=}0$); 216 ($e{=}0$) \\
X6 ca02 & enum & 40 ($e{=}0$); 84 ($e{=}0$); 312 ($e{=}0$) \\
X6 ca03 & enum & 16 ($e{=}0$); 38 ($e{=}0$); 152 ($e{=}0$) \\
X6 ca04 & enum & 20 ($e{=}0$); 43 ($e{=}0$); 208 ($e{=}0$) \\
X6 ca05 & enum & 33 ($e{=}0$); 74 ($e{=}0$); 304 ($e{=}0$) \\
\bottomrule
\end{tabular}


\end{table}

\paragraph{Row by row.}
\begin{itemize}
\item \textbf{Enumeration} (ALU, day of week, tic-tac-toe, routing; X6 routing and calendars). The certifier evaluates
every input in exact integers and records $M_i$; \cref{thm:t1} transfers the result. Routing enumerates the $2^{20}$
patterns of the 20 address bits that the weights read; all other input columns are exactly zero, and every route of
the suite and of X6 has a prefix of at most 20 bits, so the specification ignores the other bits too.
\item \textbf{DP} (Luhn; X6 check digits). The DP propagates exact integer hidden pre-activations digit by digit and
keeps the set of distinct (pre-activation vector, residue) pairs. The first layer's $M_i$ is bounded over all inputs
by $\sum_p\max_d|\hat W_{i,(p,d)}|+|\hat b_i|$, because each position has one active digit. The second layer's $M_i$
is computed over all final DP states. Both are below $2^{24}$ for all 11 networks.
\item \textbf{Template} (Sudoku, chemistry). The checker verifies the weights entry by entry against a template and
proves the template correct by counting. For Sudoku each hidden unit sees 9 cells with one digit each, so $M_i\le10$,
and the readout over 243 units in $\{0,1\}$ has $M\le2{,}435$. For chemistry the neighbour sum $A X$ adds integer bond
orders whose total is at most the valence (at most 4); the hydrogen unit has $M\le8$ and the readout $M\le4$. Masses
use non-binary constants and are not certified.
\item \textbf{SMT} (chess). z3 \citep{demoura2008z3} encodes occupancy as Booleans and every pre-activation as an
integer term (\texttt{If(occ, w, 0)} sums, Int sort), with ReLU as \texttt{If(z>0, z, 0)}. The weights are integers.
Over every 0/1 board, the convolution has $M\le76$, the sum over 64 squares $M\le128$ per unit, and the readout
$M\le78{,}085$. So the float32 network equals the integer model z3 reasons about. The convolution must be computed as
a sum of products (direct or matrix-multiply based), not by Winograd or FFT.
\item \textbf{BFS} (gene finder). The network reads DNA as one-hot bases, three per codon. The check has three parts.
(i) The codon convolution: on all 64 codons, unit $k$ outputs 1 exactly on codon $k$ and 0 otherwise. (ii) The RNN:
starting from the zero state, a breadth-first search applies all 64 codon inputs to every state it reaches. It finds 3
states and checks all 192 transitions; the state encodes ``inside a gene'' as $u_0-u_1\in\{0,1\}$. (iii) The readout:
for every (state, codon) pair the argmax must be the amino acid of the codon when inside a gene, and ``not in a gene''
otherwise. Since every reachable configuration is checked, the network is exact on every DNA string of every length
(reading frame 0). All weights are integers, and the largest $M_i$ over the reachable configurations is 10.
\item \textbf{Local} (shortest paths, maze). See the proof of \cref{thm:t2}: $M\le2B+L=8{,}197$ for the graph network
and $M\le12T$ for the maze readout.
\item \textbf{Margin} (Roman, tax). See \cref{lem:margin}. Roman: the exact error of the stored network plus the float32 bound is at most 0.0088 (value units)
against a margin of 0.5. Tax (suite): the network is within \$0.021 (single) and \$0.013 (joint) of the schedule in
exact rationals; the float32 error adds at most \$1.88 and \$1.83, for every real income in [\$0, \$2M] and every
summation order. X6 tax: exact-rational errors are \$0.001--0.012, and the certified totals are \$1.46, \$2.63,
\$1.28, \$2.10 and \$0.94 (larger of the two statuses; all bounds here are rounded up). A bound that counts all 32 units, active or not, is up to
\$19.7; leaving out units that cannot switch on, as \cref{lem:margin} allows, gives the values above.
\end{itemize}

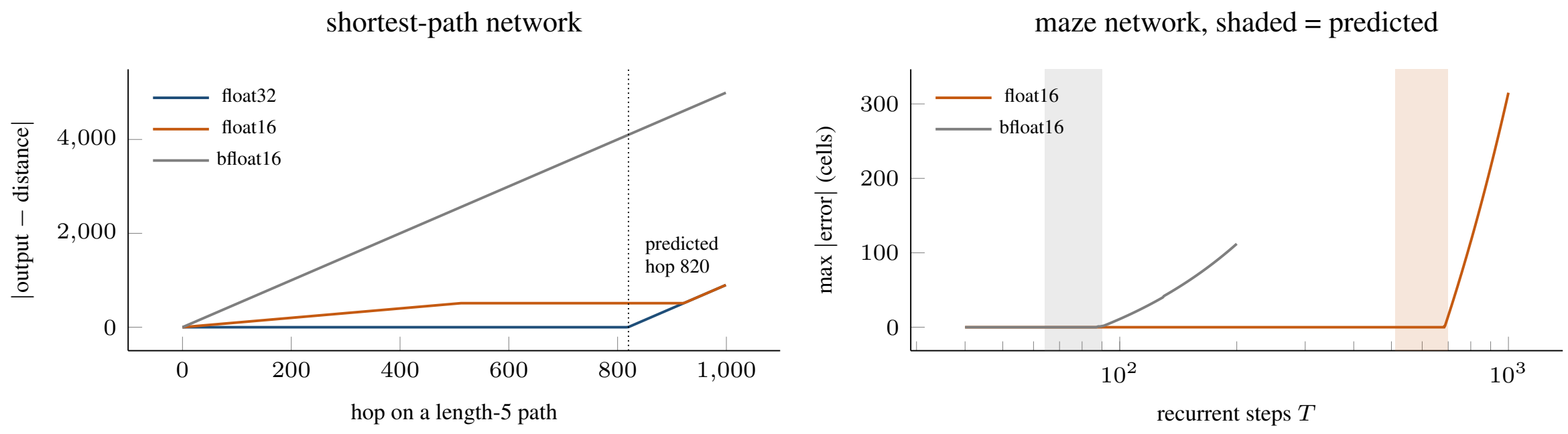
\begin{figure}[h]
\centering
\resizebox{0.9\textwidth}{!}{\begin{tikzpicture}
\definecolor{c32}{HTML}{1F4E79}\definecolor{c16}{HTML}{C55A11}\definecolor{cbf}{HTML}{7F7F7F}
\begin{axis}[name=L, width=0.5\textwidth, height=4.6cm, xlabel={hop on a length-5 path}, ylabel={$|$output $-$ distance$|$},
  title={\footnotesize shortest-path network}, tick label style={font=\scriptsize}, label style={font=\scriptsize},
  legend style={font=\tiny, draw=none, fill=none, at={(0.02,0.98)}, anchor=north west}, unbounded coords=jump,
  axis x line*=bottom, axis y line*=left, scaled y ticks=false]
\addplot[c32, line width=0.8pt] coordinates {(0,0) (4,0) (8,0) (12,0) (16,0) (20,0) (24,0) (28,0) (32,0) (36,0) (40,0) (44,0) (48,0) (52,0) (56,0) (60,0) (64,0) (68,0) (72,0) (76,0) (80,0) (84,0) (88,0) (92,0) (96,0) (100,0) (104,0) (108,0) (112,0) (116,0) (120,0) (124,0) (128,0) (132,0) (136,0) (140,0) (144,0) (148,0) (152,0) (156,0) (160,0) (164,0) (168,0) (172,0) (176,0) (180,0) (184,0) (188,0) (192,0) (196,0) (200,0) (204,0) (208,0) (212,0) (216,0) (220,0) (224,0) (228,0) (232,0) (236,0) (240,0) (244,0) (248,0) (252,0) (256,0) (260,0) (264,0) (268,0) (272,0) (276,0) (280,0) (284,0) (288,0) (292,0) (296,0) (300,0) (304,0) (308,0) (312,0) (316,0) (320,0) (324,0) (328,0) (332,0) (336,0) (340,0) (344,0) (348,0) (352,0) (356,0) (360,0) (364,0) (368,0) (372,0) (376,0) (380,0) (384,0) (388,0) (392,0) (396,0) (400,0) (404,0) (408,0) (412,0) (416,0) (420,0) (424,0) (428,0) (432,0) (436,0) (440,0) (444,0) (448,0) (452,0) (456,0) (460,0) (464,0) (468,0) (472,0) (476,0) (480,0) (484,0) (488,0) (492,0) (496,0) (500,0) (504,0) (508,0) (512,0) (516,0) (520,0) (524,0) (528,0) (532,0) (536,0) (540,0) (544,0) (548,0) (552,0) (556,0) (560,0) (564,0) (568,0) (572,0) (576,0) (580,0) (584,0) (588,0) (592,0) (596,0) (600,0) (604,0) (608,0) (612,0) (616,0) (620,0) (624,0) (628,0) (632,0) (636,0) (640,0) (644,0) (648,0) (652,0) (656,0) (660,0) (664,0) (668,0) (672,0) (676,0) (680,0) (684,0) (688,0) (692,0) (696,0) (700,0) (704,0) (708,0) (712,0) (716,0) (720,0) (724,0) (728,0) (732,0) (736,0) (740,0) (744,0) (748,0) (752,0) (756,0) (760,0) (764,0) (768,0) (772,0) (776,0) (780,0) (784,0) (788,0) (792,0) (796,0) (800,0) (804,0) (808,0) (812,0) (816,0) (820,4) (823,19) (824,24) (827,39) (828,44) (831,59) (832,64) (835,79) (836,84) (839,99) (840,104) (843,119) (844,124) (847,139) (848,144) (851,159) (852,164) (855,179) (856,184) (859,199) (860,204) (863,219) (864,224) (867,239) (868,244) (871,259) (872,264) (875,279) (876,284) (879,299) (880,304) (883,319) (884,324) (887,339) (888,344) (891,359) (892,364) (895,379) (896,384) (899,399) (900,404) (903,419) (904,424) (907,439) (908,444) (911,459) (912,464) (915,479) (916,484) (919,499) (920,504) (923,519) (924,524) (927,539) (928,544) (931,559) (932,564) (935,579) (936,584) (939,599) (940,604) (943,619) (944,624) (947,639) (948,644) (951,659) (952,664) (955,679) (956,684) (959,699) (960,704) (963,719) (964,724) (967,739) (968,744) (971,759) (972,764) (975,779) (976,784) (979,799) (980,804) (983,819) (984,824) (987,839) (988,844) (991,859) (992,864) (995,879) (996,884) (999,899)};
\addlegendentry{float32}
\addplot[c16, line width=0.8pt] coordinates {(0,0) (3,3) (4,4) (7,7) (8,8) (11,11) (12,12) (15,15) (16,16) (19,19) (20,20) (23,23) (24,24) (27,27) (28,28) (31,31) (32,32) (35,35) (36,36) (39,39) (40,40) (43,43) (44,44) (47,47) (48,48) (51,51) (52,52) (55,55) (56,56) (59,59) (60,60) (63,63) (64,64) (67,67) (68,68) (71,71) (72,72) (75,75) (76,76) (79,79) (80,80) (83,83) (84,84) (87,87) (88,88) (91,91) (92,92) (95,95) (96,96) (99,99) (100,100) (103,103) (104,104) (107,107) (108,108) (111,111) (112,112) (115,115) (116,116) (119,119) (120,120) (123,123) (124,124) (127,127) (128,128) (131,131) (132,132) (135,135) (136,136) (139,139) (140,140) (143,143) (144,144) (147,147) (148,148) (151,151) (152,152) (155,155) (156,156) (159,159) (160,160) (163,163) (164,164) (167,167) (168,168) (171,171) (172,172) (175,175) (176,176) (179,179) (180,180) (183,183) (184,184) (187,187) (188,188) (191,191) (192,192) (195,195) (196,196) (199,199) (200,200) (203,203) (204,204) (207,207) (208,208) (211,211) (212,212) (215,215) (216,216) (219,219) (220,220) (223,223) (224,224) (227,227) (228,228) (231,231) (232,232) (235,235) (236,236) (239,239) (240,240) (243,243) (244,244) (247,247) (248,248) (251,251) (252,252) (255,255) (256,256) (259,259) (260,260) (263,263) (264,264) (267,267) (268,268) (271,271) (272,272) (275,275) (276,276) (279,279) (280,280) (283,283) (284,284) (287,287) (288,288) (291,291) (292,292) (295,295) (296,296) (299,299) (300,300) (303,303) (304,304) (307,307) (308,308) (311,311) (312,312) (315,315) (316,316) (319,319) (320,320) (323,323) (324,324) (327,327) (328,328) (331,331) (332,332) (335,335) (336,336) (339,339) (340,340) (343,343) (344,344) (347,347) (348,348) (351,351) (352,352) (355,355) (356,356) (359,359) (360,360) (363,363) (364,364) (367,367) (368,368) (371,371) (372,372) (375,375) (376,376) (379,379) (380,380) (383,383) (384,384) (387,387) (388,388) (391,391) (392,392) (395,395) (396,396) (399,399) (400,400) (403,403) (404,404) (407,407) (408,408) (411,411) (412,412) (415,415) (416,416) (419,419) (420,420) (423,423) (424,424) (427,427) (428,428) (431,431) (432,432) (435,435) (436,436) (439,439) (440,440) (443,443) (444,444) (447,447) (448,448) (451,451) (452,452) (455,455) (456,456) (459,459) (460,460) (463,463) (464,464) (467,467) (468,468) (471,471) (472,472) (475,475) (476,476) (479,479) (480,480) (483,483) (484,484) (487,487) (488,488) (491,491) (492,492) (495,495) (496,496) (499,499) (500,500) (503,503) (504,504) (507,507) (508,508) (511,511) (512,512) (516,512) (520,512) (524,512) (528,512) (532,512) (536,512) (540,512) (544,512) (548,512) (552,512) (556,512) (560,512) (564,512) (568,512) (572,512) (576,512) (580,512) (584,512) (588,512) (592,512) (596,512) (600,512) (604,512) (608,512) (612,512) (616,512) (620,512) (624,512) (628,512) (632,512) (636,512) (640,512) (644,512) (648,512) (652,512) (656,512) (660,512) (664,512) (668,512) (672,512) (676,512) (680,512) (684,512) (688,512) (692,512) (696,512) (700,512) (704,512) (708,512) (712,512) (716,512) (720,512) (724,512) (728,512) (732,512) (736,512) (740,512) (744,512) (748,512) (752,512) (756,512) (760,512) (764,512) (768,512) (772,512) (776,512) (780,512) (784,512) (788,512) (792,512) (796,512) (800,512) (804,512) (808,512) (812,512) (816,512) (820,512) (824,512) (828,512) (832,512) (836,512) (840,512) (844,512) (848,512) (852,512) (856,512) (860,512) (864,512) (868,512) (872,512) (876,512) (880,512) (884,512) (888,512) (892,512) (896,512) (900,512) (904,512) (908,512) (912,512) (916,512) (920,512) (923,519) (924,524) (927,539) (928,544) (931,559) (932,564) (935,579) (936,584) (939,599) (940,604) (943,619) (944,624) (947,639) (948,644) (951,659) (952,664) (955,679) (956,684) (959,699) (960,704) (963,719) (964,724) (967,739) (968,744) (971,759) (972,764) (975,779) (976,784) (979,799) (980,804) (983,819) (984,824) (987,839) (988,844) (991,859) (992,864) (995,879) (996,884) (999,899)};
\addlegendentry{float16}
\addplot[cbf, line width=0.8pt] coordinates {(0,0) (3,15) (4,20) (7,35) (8,40) (11,55) (12,60) (15,75) (16,80) (19,95) (20,100) (23,115) (24,120) (27,135) (28,140) (31,155) (32,160) (35,175) (36,180) (39,195) (40,200) (43,215) (44,220) (47,235) (48,240) (51,255) (52,260) (55,275) (56,280) (59,295) (60,300) (63,315) (64,320) (67,335) (68,340) (71,355) (72,360) (75,375) (76,380) (79,395) (80,400) (83,415) (84,420) (87,435) (88,440) (91,455) (92,460) (95,475) (96,480) (99,495) (100,500) (103,515) (104,520) (107,535) (108,540) (111,555) (112,560) (115,575) (116,580) (119,595) (120,600) (123,615) (124,620) (127,635) (128,640) (131,655) (132,660) (135,675) (136,680) (139,695) (140,700) (143,715) (144,720) (147,735) (148,740) (151,755) (152,760) (155,775) (156,780) (159,795) (160,800) (163,815) (164,820) (167,835) (168,840) (171,855) (172,860) (175,875) (176,880) (179,895) (180,900) (183,915) (184,920) (187,935) (188,940) (191,955) (192,960) (195,975) (196,980) (199,995) (200,1000) (203,1015) (204,1020) (207,1035) (208,1040) (211,1055) (212,1060) (215,1075) (216,1080) (219,1095) (220,1100) (223,1115) (224,1120) (227,1135) (228,1140) (231,1155) (232,1160) (235,1175) (236,1180) (239,1195) (240,1200) (243,1215) (244,1220) (247,1235) (248,1240) (251,1255) (252,1260) (255,1275) (256,1280) (259,1295) (260,1300) (263,1315) (264,1320) (267,1335) (268,1340) (271,1355) (272,1360) (275,1375) (276,1380) (279,1395) (280,1400) (283,1415) (284,1420) (287,1435) (288,1440) (291,1455) (292,1460) (295,1475) (296,1480) (299,1495) (300,1500) (303,1515) (304,1520) (307,1535) (308,1540) (311,1555) (312,1560) (315,1575) (316,1580) (319,1595) (320,1600) (323,1615) (324,1620) (327,1635) (328,1640) (331,1655) (332,1660) (335,1675) (336,1680) (339,1695) (340,1700) (343,1715) (344,1720) (347,1735) (348,1740) (351,1755) (352,1760) (355,1775) (356,1780) (359,1795) (360,1800) (363,1815) (364,1820) (367,1835) (368,1840) (371,1855) (372,1860) (375,1875) (376,1880) (379,1895) (380,1900) (383,1915) (384,1920) (387,1935) (388,1940) (391,1955) (392,1960) (395,1975) (396,1980) (399,1995) (400,2000) (403,2015) (404,2020) (407,2035) (408,2040) (411,2055) (412,2060) (415,2075) (416,2080) (419,2095) (420,2100) (423,2115) (424,2120) (427,2135) (428,2140) (431,2155) (432,2160) (435,2175) (436,2180) (439,2195) (440,2200) (443,2215) (444,2220) (447,2235) (448,2240) (451,2255) (452,2260) (455,2275) (456,2280) (459,2295) (460,2300) (463,2315) (464,2320) (467,2335) (468,2340) (471,2355) (472,2360) (475,2375) (476,2380) (479,2395) (480,2400) (483,2415) (484,2420) (487,2435) (488,2440) (491,2455) (492,2460) (495,2475) (496,2480) (499,2495) (500,2500) (503,2515) (504,2520) (507,2535) (508,2540) (511,2555) (512,2560) (515,2575) (516,2580) (519,2595) (520,2600) (523,2615) (524,2620) (527,2635) (528,2640) (531,2655) (532,2660) (535,2675) (536,2680) (539,2695) (540,2700) (543,2715) (544,2720) (547,2735) (548,2740) (551,2755) (552,2760) (555,2775) (556,2780) (559,2795) (560,2800) (563,2815) (564,2820) (567,2835) (568,2840) (571,2855) (572,2860) (575,2875) (576,2880) (579,2895) (580,2900) (583,2915) (584,2920) (587,2935) (588,2940) (591,2955) (592,2960) (595,2975) (596,2980) (599,2995) (600,3000) (603,3015) (604,3020) (607,3035) (608,3040) (611,3055) (612,3060) (615,3075) (616,3080) (619,3095) (620,3100) (623,3115) (624,3120) (627,3135) (628,3140) (631,3155) (632,3160) (635,3175) (636,3180) (639,3195) (640,3200) (643,3215) (644,3220) (647,3235) (648,3240) (651,3255) (652,3260) (655,3275) (656,3280) (659,3295) (660,3300) (663,3315) (664,3320) (667,3335) (668,3340) (671,3355) (672,3360) (675,3375) (676,3380) (679,3395) (680,3400) (683,3415) (684,3420) (687,3435) (688,3440) (691,3455) (692,3460) (695,3475) (696,3480) (699,3495) (700,3500) (703,3515) (704,3520) (707,3535) (708,3540) (711,3555) (712,3560) (715,3575) (716,3580) (719,3595) (720,3600) (723,3615) (724,3620) (727,3635) (728,3640) (731,3655) (732,3660) (735,3675) (736,3680) (739,3695) (740,3700) (743,3715) (744,3720) (747,3735) (748,3740) (751,3755) (752,3760) (755,3775) (756,3780) (759,3795) (760,3800) (763,3815) (764,3820) (767,3835) (768,3840) (771,3855) (772,3860) (775,3875) (776,3880) (779,3895) (780,3900) (783,3915) (784,3920) (787,3935) (788,3940) (791,3955) (792,3960) (795,3975) (796,3980) (799,3995) (800,4000) (803,4015) (804,4020) (807,4035) (808,4040) (811,4055) (812,4060) (815,4075) (816,4080) (819,4095) (820,4100) (823,4115) (824,4120) (827,4135) (828,4140) (831,4155) (832,4160) (835,4175) (836,4180) (839,4195) (840,4200) (843,4215) (844,4220) (847,4235) (848,4240) (851,4255) (852,4260) (855,4275) (856,4280) (859,4295) (860,4300) (863,4315) (864,4320) (867,4335) (868,4340) (871,4355) (872,4360) (875,4375) (876,4380) (879,4395) (880,4400) (883,4415) (884,4420) (887,4435) (888,4440) (891,4455) (892,4460) (895,4475) (896,4480) (899,4495) (900,4500) (903,4515) (904,4520) (907,4535) (908,4540) (911,4555) (912,4560) (915,4575) (916,4580) (919,4595) (920,4600) (923,4615) (924,4620) (927,4635) (928,4640) (931,4655) (932,4660) (935,4675) (936,4680) (939,4695) (940,4700) (943,4715) (944,4720) (947,4735) (948,4740) (951,4755) (952,4760) (955,4775) (956,4780) (959,4795) (960,4800) (963,4815) (964,4820) (967,4835) (968,4840) (971,4855) (972,4860) (975,4875) (976,4880) (979,4895) (980,4900) (983,4915) (984,4920) (987,4935) (988,4940) (991,4955) (992,4960) (995,4975) (996,4980) (999,4995)};
\addlegendentry{bfloat16}
\draw[densely dotted] (axis cs:820,\pgfkeysvalueof{/pgfplots/ymin}) -- (axis cs:820,\pgfkeysvalueof{/pgfplots/ymax});
\node[font=\tiny, align=left, anchor=west] at (axis cs:830,1500) {predicted\\hop 820};
\end{axis}
\begin{axis}[at={(L.east)}, anchor=west, xshift=1.4cm, width=0.5\textwidth, height=4.6cm, xmode=log,
  xlabel={recurrent steps $T$}, ylabel={max $|$error$|$ (cells)}, title={\footnotesize maze network, shaded = predicted},
  tick label style={font=\scriptsize}, label style={font=\scriptsize}, unbounded coords=jump,
  legend style={font=\tiny, draw=none, fill=none, at={(0.02,0.98)}, anchor=north west},
  axis x line*=bottom, axis y line*=left]
\fill[c16, opacity=0.15] (axis cs:512,\pgfkeysvalueof{/pgfplots/ymin}) rectangle (axis cs:700,\pgfkeysvalueof{/pgfplots/ymax});
\fill[cbf, opacity=0.15] (axis cs:64,\pgfkeysvalueof{/pgfplots/ymin}) rectangle (axis cs:90,\pgfkeysvalueof{/pgfplots/ymax});
\addplot[c16, line width=0.8pt] coordinates {(1,nan) (3,nan) (5,nan) (7,nan) (9,nan) (11,nan) (13,nan) (15,nan) (17,nan) (19,nan) (21,nan) (23,nan) (25,nan) (27,nan) (29,nan) (31,nan) (33,nan) (35,nan) (37,nan) (40,0) (41,0) (43,0) (45,0) (47,0) (49,0) (51,0) (53,0) (55,0) (57,0) (59,0) (61,0) (63,0) (65,0) (67,0) (69,0) (71,0) (73,0) (75,0) (77,0) (79,0) (81,0) (83,0) (85,0) (87,0) (89,0) (91,0) (93,0) (95,0) (97,0) (99,0) (101,0) (103,0) (105,0) (107,0) (109,0) (111,0) (113,0) (115,0) (117,0) (119,0) (121,0) (123,0) (125,0) (127,0) (129,0) (131,0) (133,0) (135,0) (137,0) (139,0) (141,0) (143,0) (145,0) (147,0) (149,0) (151,0) (153,0) (155,0) (157,0) (159,0) (161,0) (163,0) (165,0) (167,0) (169,0) (171,0) (173,0) (175,0) (177,0) (179,0) (181,0) (183,0) (185,0) (187,0) (189,0) (191,0) (193,0) (195,0) (197,0) (199,0) (201,0) (203,0) (205,0) (207,0) (209,0) (211,0) (213,0) (215,0) (217,0) (219,0) (221,0) (223,0) (225,0) (227,0) (229,0) (231,0) (233,0) (235,0) (237,0) (239,0) (241,0) (243,0) (245,0) (247,0) (249,0) (251,0) (253,0) (255,0) (257,0) (259,0) (261,0) (263,0) (265,0) (267,0) (269,0) (271,0) (273,0) (275,0) (277,0) (279,0) (281,0) (283,0) (285,0) (287,0) (289,0) (291,0) (293,0) (295,0) (297,0) (299,0) (301,0) (303,0) (305,0) (307,0) (309,0) (311,0) (313,0) (315,0) (317,0) (319,0) (321,0) (323,0) (325,0) (327,0) (329,0) (331,0) (333,0) (335,0) (337,0) (339,0) (341,0) (343,0) (345,0) (347,0) (349,0) (351,0) (353,0) (355,0) (357,0) (359,0) (361,0) (363,0) (365,0) (367,0) (369,0) (371,0) (373,0) (375,0) (377,0) (379,0) (381,0) (383,0) (385,0) (387,0) (389,0) (391,0) (393,0) (395,0) (397,0) (399,0) (401,0) (403,0) (405,0) (407,0) (409,0) (411,0) (413,0) (415,0) (417,0) (419,0) (421,0) (423,0) (425,0) (427,0) (429,0) (431,0) (433,0) (435,0) (437,0) (439,0) (441,0) (443,0) (445,0) (447,0) (449,0) (451,0) (453,0) (455,0) (457,0) (459,0) (461,0) (463,0) (465,0) (467,0) (469,0) (471,0) (473,0) (475,0) (477,0) (479,0) (481,0) (483,0) (485,0) (487,0) (489,0) (491,0) (493,0) (495,0) (497,0) (499,0) (501,0) (503,0) (505,0) (507,0) (509,0) (511,0) (513,0) (515,0) (517,0) (519,0) (521,0) (523,0) (525,0) (527,0) (529,0) (531,0) (533,0) (535,0) (537,0) (539,0) (541,0) (543,0) (545,0) (547,0) (549,0) (551,0) (553,0) (555,0) (557,0) (559,0) (561,0) (563,0) (565,0) (567,0) (569,0) (571,0) (573,0) (575,0) (577,0) (579,0) (581,0) (583,0) (585,0) (587,0) (589,0) (591,0) (593,0) (595,0) (597,0) (599,0) (601,0) (603,0) (605,0) (607,0) (609,0) (611,0) (613,0) (615,0) (617,0) (619,0) (621,0) (623,0) (625,0) (627,0) (629,0) (631,0) (633,0) (635,0) (637,0) (639,0) (641,0) (643,0) (645,0) (647,0) (649,0) (651,0) (653,0) (655,0) (657,0) (659,0) (661,0) (663,0) (665,0) (667,0) (669,0) (671,0) (673,0) (675,0) (677,0) (679,0) (681,0) (683,0) (684,1) (685,1) (687,2) (688,3) (689,4) (690,5) (691,6) (692,7) (693,8) (694,9) (695,10) (696,11) (697,12) (698,13) (699,14) (700,15) (701,16) (702,17) (703,18) (704,19) (705,20) (706,21) (707,22) (708,23) (709,24) (710,25) (711,26) (712,27) (713,28) (714,29) (715,30) (716,31) (717,32) (718,33) (719,34) (720,35) (721,36) (722,37) (723,38) (724,39) (725,40) (726,41) (727,42) (728,43) (729,44) (730,45) (731,46) (732,47) (733,48) (734,49) (735,50) (736,51) (737,52) (738,53) (739,54) (740,55) (741,56) (742,57) (743,58) (744,59) (745,60) (746,61) (747,62) (748,63) (749,64) (750,65) (751,66) (752,67) (753,68) (754,69) (755,70) (756,71) (757,72) (758,73) (759,74) (760,75) (761,76) (762,77) (763,78) (764,79) (765,80) (766,81) (767,82) (768,83) (769,84) (770,85) (771,86) (772,87) (773,88) (774,89) (775,90) (776,91) (777,92) (778,93) (779,94) (780,95) (781,96) (782,97) (783,98) (784,99) (785,100) (786,101) (787,102) (788,103) (789,104) (790,105) (791,106) (792,107) (793,108) (794,109) (795,110) (796,111) (797,112) (798,113) (799,114) (800,115) (801,116) (802,117) (803,118) (804,119) (805,120) (806,121) (807,122) (808,123) (809,124) (810,125) (811,126) (812,127) (813,128) (814,129) (815,130) (816,131) (817,132) (818,133) (819,134) (820,135) (821,136) (822,137) (823,138) (824,139) (825,140) (826,141) (827,142) (828,143) (829,144) (830,145) (831,146) (832,147) (833,148) (834,149) (835,150) (836,151) (837,152) (838,153) (839,154) (840,155) (841,156) (842,157) (843,158) (844,159) (845,160) (846,161) (847,162) (848,163) (849,164) (850,165) (851,166) (852,167) (853,168) (854,169) (855,170) (856,171) (857,172) (858,173) (859,174) (860,175) (861,176) (862,177) (863,178) (864,179) (865,180) (866,181) (867,182) (868,183) (869,184) (870,185) (871,186) (872,187) (873,188) (874,189) (875,190) (876,191) (877,192) (878,193) (879,194) (880,195) (881,196) (882,197) (883,198) (884,199) (885,200) (886,201) (887,202) (888,203) (889,204) (890,205) (891,206) (892,207) (893,208) (894,209) (895,210) (896,211) (897,212) (898,213) (899,214) (900,215) (901,216) (902,217) (903,218) (904,219) (905,220) (906,221) (907,222) (908,223) (909,224) (910,225) (911,226) (912,227) (913,228) (914,229) (915,230) (916,231) (917,232) (918,233) (919,234) (920,235) (921,236) (922,237) (923,238) (924,239) (925,240) (926,241) (927,242) (928,243) (929,244) (930,245) (931,246) (932,247) (933,248) (934,249) (935,250) (936,251) (937,252) (938,253) (939,254) (940,255) (941,256) (942,257) (943,258) (944,259) (945,260) (946,261) (947,262) (948,263) (949,264) (950,265) (951,266) (952,267) (953,268) (954,269) (955,270) (956,271) (957,272) (958,273) (959,274) (960,275) (961,276) (962,277) (963,278) (964,279) (965,280) (966,281) (967,282) (968,283) (969,284) (970,285) (971,286) (972,287) (973,288) (974,289) (975,290) (976,291) (977,292) (978,293) (979,294) (980,295) (981,296) (982,297) (983,298) (984,299) (985,300) (986,301) (987,302) (988,303) (989,304) (990,305) (991,306) (992,307) (993,308) (994,309) (995,310) (996,311) (997,312) (998,313) (999,314) (1000,315)};
\addlegendentry{float16}
\addplot[cbf, line width=0.8pt] coordinates {(1,nan) (3,nan) (5,nan) (7,nan) (9,nan) (11,nan) (13,nan) (15,nan) (17,nan) (19,nan) (21,nan) (23,nan) (25,nan) (27,nan) (29,nan) (31,nan) (33,nan) (35,nan) (37,nan) (40,0) (41,0) (43,0) (45,0) (47,0) (49,0) (51,0) (53,0) (55,0) (57,0) (59,0) (61,0) (63,0) (65,0) (67,0) (69,0) (71,0) (73,0) (75,0) (77,0) (79,0) (81,0) (83,0) (85,0) (87,0) (88,1) (89,1) (91,2) (92,3) (93,4) (94,5) (95,6) (96,7) (97,8) (98,9) (99,10) (100,11) (101,12) (102,13) (103,14) (104,15) (105,16) (106,17) (107,18) (108,19) (109,20) (110,21) (111,22) (112,23) (113,24) (114,25) (115,26) (116,27) (117,28) (118,29) (119,30) (120,31) (121,32) (122,33) (123,34) (124,35) (125,36) (126,37) (127,38) (128,39) (129,40) (130,42) (131,43) (132,44) (133,45) (134,46) (135,47) (136,48) (137,49) (138,50) (139,51) (140,52) (141,53) (142,54) (143,55) (144,56) (145,57) (146,58) (147,59) (148,60) (149,61) (150,62) (151,63) (152,64) (153,65) (154,66) (155,67) (156,68) (157,69) (158,70) (159,71) (160,72) (161,73) (162,74) (163,75) (164,76) (165,77) (166,78) (167,79) (168,80) (169,81) (170,82) (171,83) (172,84) (173,85) (174,86) (175,87) (176,88) (177,89) (178,90) (179,91) (180,92) (181,93) (182,94) (183,95) (184,96) (185,97) (186,98) (187,99) (188,100) (189,101) (190,102) (191,103) (192,104) (193,105) (194,106) (195,107) (196,108) (197,109) (198,110) (199,111) (200,112)};
\addlegendentry{bfloat16}
\end{axis}
\end{tikzpicture}}
\caption{Left: the shortest-path network is exact through hop 819 and clamps at 4096 from hop 820 in float32; in
16-bit formats it is wrong from hop 1. Right: the maze network on a 40-cell corridor first errs at $T=684$ (float16)
and $T=88$ (bfloat16).}
\label{fig:x3}
\end{figure}

\paragraph{Maze in 16-bit formats.} We ran the corridor of \cref{fig:x3} in exact integers for $T\le3{,}000$ and
recorded every stored value (the input map, the recurrent sum, the state, the accumulators and the readout). A
format with a $p$-bit significand holds an integer $N$ exactly iff $|N|\le2^p$ or $N$ is a multiple of
$2^{\lfloor\log_2|N|\rfloor-p+1}$, and $|N|$ is at most the largest finite value (65,504 for float16). The first step with a value that float16 ($p=11$) cannot hold is $T=684$; for
bfloat16 ($p=8$) it is $T=88$. These are exactly the measured failure points. On the corridor the goal accumulator
grows by 3 per step, not by 4 as the logged prediction assumed.


\section{GD baselines: budgets, grids and test sizes}
\label{app:gd}

\paragraph{Protocol.} Grid: learning rate $\{3\times10^{-4},10^{-3},3\times10^{-3}\}\times$ width
$\{\times1,\times4\}\times$ weight decay $\{0$, AdamW $10^{-2}\}$, cosine decay, gradient clipping at 1, seed 0. Each
configuration is scored on a validation split from the training distribution: fresh draws (Luhn, Sudoku, chemistry);
7,000 held-out dates of 1900--2099 (day of week); 200 or 99 held-out numerals (Roman); 10\% of the 1\% training split
(ALU); 200 fresh $10\times10$ mazes (maze); 10\% of the fixed training set (routing, chess, Hamming). A curriculum
variant of the best configuration was added for Luhn (2, 4, 8, 16 random trailing digits) and the maze ($5\to7\to10$).
Chess used width $\times1$ only, for CPU cost. Selection runs are shorter than final runs for most tasks
(\cref{tab:gdtest}, last column); for Sudoku this hid the final result, since the selection score after 8,000 steps
was 60\% and the final runs reached 100\%. Training: Adam (AdamW when decayed, on all parameters including biases), batch 32--256, PyTorch default
initialisation; losses: cross-entropy (Luhn mix, Sudoku, day of week, routing, chess), binary cross-entropy (ALU,
Hamming), smooth L1 (Roman, maze) and squared error (Luhn uniform, chemistry). Sudoku's ``other 4 kinds'' are one
changed cell, a row swap, a box swap and random rows. \Cref{tab:grid} gives every grid cell's validation score;
\cref{tab:gdtest} gives the test sizes, per-seed scores and per-seed 95\% Clopper--Pearson intervals. Code:
\texttt{code/x5\_gd.py}, \texttt{code/x10\_gd\_more.py}.

\begin{table}[h]
\caption{Validation score (\%; AUC for Luhn uniform) of every grid cell, seed 0. Columns: learning rate, width,
weight decay (-- none, wd AdamW $10^{-2}$); last column: curriculum variant. Bold: selected.}
\label{tab:grid}
\centering\scriptsize
\setlength{\tabcolsep}{2.5pt}
\begin{tabular}{@{}lrrrrrrrrrrrrr@{}}
\toprule
 & 0.0003 & 0.0003 & 0.0003 & 0.0003 & 0.001 & 0.001 & 0.001 & 0.001 & 0.003 & 0.003 & 0.003 & 0.003 & \\
 & $\times$1 & $\times$1 & $\times$4 & $\times$4 & $\times$1 & $\times$1 & $\times$4 & $\times$4 & $\times$1 & $\times$1 & $\times$4 & $\times$4 & \\
task & -- & wd & -- & wd & -- & wd & -- & wd & -- & wd & -- & wd & cur. \\
\midrule
Luhn (mix) & 50.1 & 50.0 & \textbf{51.0} & 50.2 & 50.7 & 50.1 & 50.2 & 50.2 & 50.0 & 49.7 & 49.4 & 50.8 & 50.0 \\
Luhn (uniform, AUC) & 0.500 & 0.501 & 0.498 & 0.498 & 0.498 & \textbf{0.505} & 0.497 & 0.503 & 0.503 & 0.502 & 0.498 & 0.494 & -- \\
Sudoku & 57.4 & 57.6 & 59.6 & \textbf{60.1} & 52.5 & 53.5 & 53.7 & 54.5 & 52.7 & 51.8 & 52.3 & 53.0 & -- \\
day of week & 89.6 & 87.8 & 98.4 & 97.9 & 99.4 & 99.3 & 99.9 & 99.9 & \textbf{100.0} & 99.9 & 100.0 & 99.9 & -- \\
day of week (scalar) & 82.0 & 81.9 & 97.0 & 96.9 & 99.2 & 99.5 & 99.7 & 99.8 & 99.9 & 99.9 & \textbf{100.0} & 100.0 & -- \\
Roman (random) & 95.5 & 95.5 & 99.0 & 98.5 & \textbf{100.0} & 100.0 & 97.0 & 97.0 & 97.0 & 97.0 & 100.0 & 100.0 & -- \\
Roman (1--999) & 99.0 & 99.0 & 99.0 & 99.0 & 98.0 & 99.0 & 99.0 & 99.0 & 99.0 & 99.0 & \textbf{100.0} & 100.0 & -- \\
ALU 1\% & 66.4 & 65.4 & 74.0 & 76.1 & 75.2 & 79.2 & 75.8 & 74.0 & 74.3 & \textbf{81.0} & 78.3 & 77.7 & -- \\
maze & 21.2 & 21.2 & 30.6 & 30.6 & 26.1 & 26.3 & 54.5 & 54.4 & 34.1 & 34.2 & 94.6 & 92.4 & \textbf{97.2} \\
routing & 94.0 & 94.0 & 97.0 & 97.0 & 97.3 & 97.3 & \textbf{98.3} & 98.3 & 98.0 & 98.0 & 98.3 & 98.3 & -- \\
chess & 88.6 & 88.5 &  &  & \textbf{89.5} & 89.2 &  &  & 89.1 & 88.2 &  &  & -- \\
Hamming & 94.2 & 94.2 & \textbf{97.7} & 97.7 & 94.0 & 94.1 & 97.7 & 97.7 & 95.7 & 95.7 & 97.6 & 97.6 & -- \\
chemistry & 98.9 & 98.9 & 99.4 & 99.4 & 99.6 & 99.6 & 99.9 & 99.9 & 99.9 & 99.9 & 100.0 & \textbf{100.0} & -- \\
\bottomrule
\end{tabular}


\end{table}

\begin{table}[h]
\caption{Test sets, per-seed scores and 95\% intervals of the final runs. The seeds share one test set, so instead of pooling we give the range of the three per-seed 95\% Clopper--Pearson intervals (lowest lower end to highest upper end). ``all'': the test is every input of its domain, so there is no sampling noise. Maze cells within a maze are not independent, so no interval is given; Hamming bit errors cluster within words, so its SE is approximate. Select ``--'': a fixed configuration (the written width), not selected.}
\label{tab:gdtest}
\centering\scriptsize
\setlength{\tabcolsep}{3pt}
\begin{tabular}{@{}llrllll@{}}
\toprule
task & test & $n$ per seed & seed 0 / 1 / 2 (\%) & 95\% CI, per seed (\%) & select / final steps \\
\midrule
Luhn (mix) & valid numbers & 20,000 & 49.87 / 50.03 / 49.32 & 48.62--50.73 & 20k / 40k \\
Luhn (mix) & invalid: random & 20,000 & 49.72 / 50.41 / 49.54 & 48.85--51.11 & 20k / 40k \\
Luhn (mix) & invalid: one digit & 20,000 & 50.33 / 50.77 / 49.71 & 49.01--51.46 & 20k / 40k \\
Luhn (mix) & invalid: swap & 20,000 & 50.26 / 51.02 / 49.61 & 48.91--51.72 & 20k / 40k \\
Luhn (uniform) & AUC, 200k strings & 200,000 & 0.5019 / 0.4992 / 0.5028 & SE $\approx0.002$ & 10k / 40k \\
day of week & other centuries & 219,145 & 7.16 / 6.73 / 15.38 & -- (all) & 15k / 30k \\
day of week (scalar) & other centuries & 219,145 & 1.22 / 0.64 / 2.87 & -- (all) & 15k / 30k \\
Roman & random split & 1,999 & 97.15 / 97.25 / 95.80 & -- (all) & 15k / 15k \\
Roman & 1000--3999 & 3,000 & 0.00 / 0.00 / 0.00 & -- (all) & 15k / 15k \\
Sudoku ($\times4$) & valid grids & 2,000 & 100.00 / 100.00 / 100.00 & 99.82--100.00 & 8k / 30k \\
Sudoku ($\times4$) & invalid: column swap & 2,000 & 100.00 / 100.00 / 100.00 & 99.82--100.00 & 8k / 30k \\
Sudoku ($\times4$) & invalid: other 4 kinds & 8,000 & 100.00 / 100.00 / 100.00 & 99.95--100.00 & 8k / 30k \\
Sudoku ($\times1$) & valid grids & 2,000 & 100.00 / 100.00 / 100.00 & 99.82--100.00 & -- / 30k \\
Sudoku ($\times1$) & invalid: column swap & 2,000 & 100.00 / 100.00 / 100.00 & 99.82--100.00 & -- / 30k \\
Sudoku ($\times1$) & invalid: other 4 kinds & 8,000 & 100.00 / 100.00 / 100.00 & 99.95--100.00 & -- / 30k \\
ALU 1\% & ADD, unseen inputs & 64,904 & 46.31 / 54.47 / 42.16 & -- (all) & 20k / 30k \\
ALU 1\% & SUB, unseen inputs & 64,894 & 47.80 / 51.52 / 44.34 & -- (all) & 20k / 30k \\
maze ($\times4$, cur.) & $10\times10$ cells & 500 mazes & 99.692 / 99.336 / 98.589 & -- & 3k / 6k \\
maze ($\times4$, cur.) & $30\times30$ cells & 100 mazes & 0.146 / 0.005 / 0.005 & -- & 3k / 6k \\
maze ($\times1$, cur.) & $10\times10$ cells & 500 mazes & 24.990 / 49.805 / 51.624 & -- & -- / 6k \\
routing & 200k packets & 200,000 & 98.22 / 98.46 / 98.31 & 98.16--98.51 & 10k / 20k \\
chess & 20k positions & 20,000 & 88.88 / 89.14 / 89.06 & 88.44--89.57 & 3k / 15k \\
Hamming & BER at 4\,dB & $4\times10^6$ bits & 0.646 / 0.619 / 0.630 & SE $\approx0.004$ & 6k / 6k \\
chemistry & H counts, 100 atoms & 1,000 mol. & 100.00 / 100.00 / 100.00 & 99.63--100.00 & 8k / 8k \\
chemistry & mass MAE, 100 atoms (Da) & 1,000 mol. & 0.174 / 0.223 / 0.311 & -- & 8k / 8k \\
\bottomrule
\end{tabular}


\end{table}

\paragraph{Tax extrapolation (stored networks, not tuned).} The nine GD tax networks stored in the notebook were
trained on incomes up to \$500k. Above \$500k every one under-predicts, and its shortfall correlates at $\ge0.99$ with
the unseen bracket increases. \Cref{thm:t3}(b) only says that at least one of two targets that agree on the training range is extrapolated
wrongly; it says nothing about this schedule or the sign. We inspected the weights (\texttt{code/x5\_tax\_hinges.py}):
in 15 of the 18 network--status pairs, 2--7 hidden units that are off in training switch on between \$500k and
\$2M, but they change the slope by only $-4.7$ to $+8$ points. The marginal rate on \$1.5M--2M is 30.3--37.0\%,
against a true 37\% above \$624k (single) or \$760k (joint) gross income, so the networks fall behind the rising schedule. Our size prediction failed: we predicted a mean error of
\$22.5k and measured \$30.8k (range \$3.5k--43.1k), because the single-filer networks also under-fit the last in-range
slope (30--34\% instead of 35\%). These baselines were not tuned, so \cref{tab:evidence} lists tax as theorem only.

\section{Writer prompt and protocol}
\label{app:prompt}

Each of the 30 sessions started with one line: ``Read exactly ONE file, \texttt{<path to its prompt>}, and carry out
the task it describes exactly as written. Reading that single file is the only exception to its `do not read any file'
rule; do not read, list or search anything else on disk.'' The prompt for \texttt{cd01} follows. We replaced local
paths with placeholders in angle brackets; nothing else is changed. The other families differ only in the
specification and network paragraphs.
\begingroup\tiny
\begin{verbatim}
You are a fresh writer session in a study of LLM weight-writing. Write, by hand (no training, no optimisation, no
gradient steps, no fitting), the weights of a small fixed neural network that computes the specification below EXACTLY.

THE SPECIFICATION (a check-digit scheme; it is new and has no standard name):
A string of n = 8 decimal digits d_0 d_1 ... d_7 (position 0 is the leftmost) is VALID iff
    sum over p of  a_p * g_p(d_p)   is congruent to 0 modulo m = 13,
where for each position p, a_p is an integer and g_p is a table: g_p(d) = g[d] (the d-th entry of the list g).
  position 0: a = 7, g = [2, 8, 3, 1, 6, 9, 0, 7, 4, 5]
  position 1: a = 3, g = [6, 1, 4, 2, 0, 3, 7, 8, 5, 9]
  position 2: a = 11, g = [7, 2, 5, 1, 6, 8, 3, 0, 4, 9]
  position 3: a = 1, g = [0, 2, 8, 9, 4, 5, 6, 7, 3, 1]
  position 4: a = 7, g = [0, 4, 8, 3, 5, 7, 1, 2, 9, 6]
  position 5: a = 6, g = [6, 9, 5, 8, 1, 2, 3, 4, 0, 7]
  position 6: a = 1, g = [0, 4, 9, 3, 2, 1, 7, 5, 8, 6]
  position 7: a = 9, g = [0, 2, 8, 6, 4, 3, 5, 7, 1, 9]
Every one of the 10^8 digit strings counts; the checker covers all of them.

THE NETWORK (fixed; you may not change it): input x = concatenation of the one-hot codes of the 8 digits
(x[10*p + d] = 1 iff digit p equals d; 80 inputs). Then
  h = ReLU(fc1.weight @ x + fc1.bias)       (256 hidden units)
  logits = fc2.weight @ h + fc2.bias        (2 logits)
prediction = argmax(logits); class 1 = VALID, class 0 = invalid. A tie between the two logits counts as wrong.
Shapes: fc1.weight [256, 80], fc1.bias [256], fc2.weight [2, 256], fc2.bias [2]. Forward pass in float32.

HOW TO SAVE: create the file <OUT_DIR>\cd01.py with this skeleton and run it once with
<PYTHON>:
  import numpy as np
  W = {
    "fc1.weight": np.zeros((256, 80), np.float32),
    "fc1.bias": np.zeros((256,), np.float32),
    "fc2.weight": np.zeros((2, 256), np.float32),
    "fc2.bias": np.zeros((2,), np.float32),
  }
  ... set the entries of W ...
  np.savez(r"<OUT_DIR>\cd01.npz", **W)
Comment the construction in the file.

RULES (strict; they define the experiment):
- Do NOT read, list or search any file or folder under <PROJECT_ROOT> (or anywhere else on disk). You only need this message.
- Do NOT evaluate the network before submission: do not run its forward pass on any input, do not simulate it,
  do not compare its outputs with the specification, do not write tests. Do NOT fit, optimise, regress or search
  for weights. Python may be used only to build and save the weights; you may use Python arithmetic to compute
  constants or tables directly from the specification (for example a per-digit table or an offset).
- Write one file and run it once. Your first saved submission is final and is scored once by an exhaustive checker.
When done, reply with: the construction in 3-6 sentences, the number of hidden units used, the number of non-zero
weights, and a list of every tool call you made (tool name and purpose).
\end{verbatim}
\endgroup

\paragraph{Generator.} \texttt{code/x6\_specs.py}, seed 20261123. Check digits: $m\sim U\{7..13\}$,
$n\sim U\{8..14\}$, multipliers uniform among units mod $m$, each $g_p$ a uniform permutation of $0..9$. Routing:
$\max(4,\lfloor r/4\rfloor)$ base prefixes of 4--12 bits, then extensions by 1--5 bits up to 20 bits, $r\sim
U\{16..48\}$ routes, ports uniform with the extension's port different from its parent's. Tax: 4--8 brackets (4--6
with a credit), rates distinct integers in 5--45\%, thresholds multiples of \$50, joint thresholds 1.5, 1.75 or 2
times single. Calendars: week length 7 for \texttt{ca01} and uniform in $\{5,6,8,9,10\}$ otherwise, month lengths in
28--31 summing to 365, leap day in a random month of at most 30 days, 1--4 exceptional centuries.

\paragraph{Order of events (file times, 2026-11-23, local time).} Specifications and prompts 14:08; scorer, including
the tax rule, 14:09; reference solutions 14:09, all 30 passing the scorer at 14:11; writer submissions 14:12--14:15.
The generator, the prompt template, the scorer and the reference solutions were written by the author session that
ran the study (the same model family, with a human directing the programme).

\section{Session context and process audit}
\label{app:audit}

We read the 30 writer logs (Claude Code CLI 2.1.330, general-purpose sub-agents) without changing them.
\begin{itemize}
\item \textbf{Model.} The model field is \texttt{claude-opus-5-5} in all 30. The harness's model identity, which each
session receives, states a knowledge cutoff of June 2026.
\item \textbf{Fixed context.} The harness system prompt for a general-purpose agent; tool definitions and the names
of deferred tools; a list of installed skills with one-line descriptions (about 23k characters); a list of agent
types; the platform and working directory; the date; the model identity; and the project's auto-memory index. The
index (\texttt{MEMORY.md}) has 14 one-line entries, such as ``Lab notebook practice'', ``WSP MNIST study'', ``Test set
only for final scoring'', ``Thesis topic proposal'' and ``Paper portfolio''. It names no task, construction or
weight. No project instruction file (\texttt{CLAUDE.md}) existed. None of the files the index links to was loaded. The
context was identical in all 30 sessions.
\item \textbf{Tool calls.} 13 sessions: Read (own prompt), Write (script), Bash (run it). 8 sessions: Read, Write,
PowerShell. 9 sessions: Read, then one Bash call that wrote and ran the script. No session read another file, listed a
folder or searched.
\item \textbf{Programs.} The 30 saved scripts contain no forward pass, \texttt{np.maximum}, matrix product,
\texttt{argmax}, random draw or calendar library call outside comments (a pattern search). The harness did not block
Python, so compliance rests on this audit. The sessions' own lists of tool calls agree with the logs.
\end{itemize}

\section{Suite origin and correction history}
\label{app:origin}

\paragraph{Origin and inclusion rule.} The notebook (2026-11-11 to 2026-11-14) has a series of single-session
domain tasks. We took all 12 networks of its last domain batch (gene finder, Hamming, Sudoku, chemistry, Roman, ALU,
tax, chess, Luhn, maze, routing, day of week) and three earlier non-transformer networks whose task has an exact or
closed-form Bayes specification (shortest paths, tic-tac-toe, motion). We left out transformers (covered by compiled
transformer work), formal-language RNNs, networks without such a specification (for example a
telephone-tone decoder and a hand-written transformer language model), and the Game of Life networks (1 and 4 generations), which would qualify. We certified them afterwards: on all
512 binary $3\times3$ neighbourhoods each block outputs exactly the Life rule, and $\max M_i=44$, so both are exact
in float32 on every torus board of side $\ge3$ (\texttt{code/x11\_life.py}). We keep the suite as first defined.
The 15 networks are case studies, not a sample from which to estimate a rate. Each network was written by one session from a specification and a fixed architecture,
with no data. The notebook records whether the first submission was exact; this record is not independently
verifiable, and we treat the 15 networks as given artefacts whose exactness we certify here.

\paragraph{Corrections to the notebook.} We corrected these notebook claims before reusing the networks. (i) The 2024
tax figures had been written from memory; all 14 thresholds and both standard deductions match the IRS announcement
\citep{irs2024}. The task is the 2024 Tax Rate Schedule for single and joint filers with the standard deduction only,
not the IRS Tax Table used below \$100k taxable income. (ii) Day of week is ``exact after one constant fix''. (iii)
The Hamming decoder was called provably optimal; it matches block ML except near ties, not the bit-optimal decoder
(\cref{prop:ham}); the planning note's gain $2/\sigma^2$ is also off by a factor 2. (iv) The chess detector had been
tested on random placements, many of them illegal. We re-test on 100,132 legal positions from 663 random self-play
games with python-chess \citep{pythonchess} (2.76\% checks; 0 errors) and prove it correct with z3. (v) ``Exact at all
sizes'' for shortest paths is replaced by \cref{thm:t2}'s bound. (vi) The claim that Sudoku is a rule GD cannot find
is withdrawn. (vii) The explanation of the Roman GD failure is corrected: the letter M does occur in training, inside
CM.

\section{Logged predictions and outcomes}
\label{app:preds}

\Cref{tab:predictions} lists every prediction in the log, in order. Basis: \emph{theorem} if it follows from a
theorem or exact algebra; \emph{check (T1 premise)} if it verifies a theorem's hypothesis; \emph{theory-motivated}
if a theorem suggests it but does not imply it; \emph{heuristic} if it is a rough estimate; \emph{empirical} otherwise, including \emph{empirical (L1 tightness)}
for how tight a bound turns out.

\begin{table}[h]
\caption{Predictions logged before each run (local logs, self-attested) and outcomes.}
\label{tab:predictions}
\centering\scriptsize
\setlength{\tabcolsep}{3pt}
\begin{tabular}{@{}lllp{4.6cm}@{}}
\toprule
exp. & prediction & basis & held? \\
\midrule
X1 & ALU, day of week, Roman, tic-tac-toe, routing exact; 16 Hamming codewords decode & empirical & yes \\
X1 & day of week integer magnitudes $<2^{10}$ & empirical & no ($1.3\times10^6$) \\
X1 & routing: a few hundred trie cells & empirical & no (89 cells) \\
X1 & tic-tac-toe: 0 losses in 549 games & empirical & in part: 33 tied positions not foreseen; 0 losses over every tie-break \\
X2 & Luhn, Sudoku, tax, chess, chemistry hydrogens certify & empirical & yes \\
X2 & chemistry mass error $<0.001$\,Da per 100 atoms & empirical & no result (mass not certified) \\
X2 & tax gate errors start at \$3,029,200 / \$3,014,600 & theorem (algebra) & yes \\
X2 & stored GD chess nets: recall drops on legal positions & empirical & yes (87.5--88.2\%) \\
X3 & shortest path: hops 820 and 4097; 16-bit wrong from hop 1 & theorem (T2) & yes \\
X3 & maze bands fp16 $[512,700]$, bf16 $[64,90]$ & heuristic & yes (684, 88); lower ends used a wrong growth rate \\
X4 & maze: 128 local configurations agree & theorem (T2 form) & in part: 80 configurations (48 excluded by the invariant) \\
X4 & shortest-path update $=\max$ on 16.8M pairs & theorem (T2) & yes \\
X4 & gene finder $\le3$ states & empirical (BFS) & yes \\
X5 & Luhn at chance on every kind & theory-motivated (T4) & yes \\
X5 & day of week $<20\%$ on unseen centuries & theory-motivated (T3a) & yes \\
X5 & Roman: random split $<100\%$; 0\% on $\ge1000$ & empirical & yes \\
X5 & ALU ADD/SUB $<90\%$ at 1\% data & empirical & yes \\
X5 & maze: $\ge80\%$ at $10\times10$, $<20\%$ at $30\times30$ & empirical & yes \\
X5 & $\ge1$ Sudoku seed near chance & empirical & \textbf{no} (all 100\%) \\
X5 & tax extrapolation error $\approx\$22.5$k & theory-motivated (T3b) & direction yes; size no (\$30.8k) \\
X5b & Sudoku $\times1$: $\ge2/3$ seeds $\ge99\%$ & empirical & yes (3/3) \\
X5b & maze $\times1$: 40--70\% in size, $<5\%$ at 30 & empirical & no (25--52\%, $\le9\%$) \\
X6 & $\ge80\%$ first-try exact & empirical & yes (25/30 certified; 30/30 pass the rule) \\
X6 & failures in derived constants & empirical & untestable (no failures) \\
X7 & Hamming $\approx$ block ML; motion $\approx$ Bayes & theorem (Props) & yes \\
X8 & random tests miss $\ge1\%$ of breaking mutants & empirical & pooled yes; ALU, day of week no \\
theory & Adam leaves the Roman M column at init (implicit) & theorem (misapplied) & no: M occurs in CM \\
theory & T3(a) day-of-week columns stay at init; T4(b), T4(c) checks & theorem & yes \\
\midrule
X8b & uniform Luhn test misses $\ge11$ of 51; pooled $\ge7\%$ & empirical & yes (12 of 51; pooled 7.4\%) \\
X9 & every T1 row has $\max M_i\le2^{24}$ & check (T1 premise) & yes \\
X9 & Roman bound below margin by $\ge100\times$ & empirical (L1 tightness) & no ($57\times$) \\
X9 & suite tax bound $\le\$0.50$ & empirical (L1 tightness) & no (\$1.90) \\
X9 & X6 tax bound $\le\$1$ for all 5 & empirical (L1 tightness) & no (1 of 5) \\
X9 & X6 tax exact to the cent: 0/5 & empirical & yes \\
X10 & Luhn, uniform digits: AUC in $[0.49,0.51]$ & theory-motivated (T4) & yes (0.499--0.503) \\
X11 & day of week, scalar century: unseen $<25\%$ & theory-motivated (T3b) & yes (0.6--2.9\%) \\
X12 & routing 98.5--99.95\% & empirical & no (98.2--98.5\%) \\
X12 & chess 90--99\%, never error-free & empirical & no (88.9--89.1\%; never error-free held) \\
X12 & Hamming 4\,dB: best $\le0.60\%$, worst $\le0.80\%$ & empirical & in part (best 0.619\%) \\
X12 & chemistry: hydrogens 100\%; mass $>0.01$\,Da at 100 atoms & empirical & yes \\
Qwen-1 & basic harness: 10--22 of 30; gap to Claude $p<0.05$ & empirical & no (23); yes ($p=0.016$) \\
Qwen-1 & routing $\ge6$/10, check digit 3--8/10, tax $\ge3$/5, calendar $\le2$/5 & empirical & yes, no (9), yes, yes \\
Qwen-2 & expanded setup: 24--28 of 30; calendar 1--3/5 & empirical & yes (25; 1/5) \\
Qwen-2 & gap to Claude $p>0.05$; $\le3$ sessions at the 40k cap & empirical & yes ($p=0.063$); no (8) \\
\bottomrule
\end{tabular}
\end{table}

\section{Numerical checks of the theory and mutation testing}
\label{app:checks}

All values from \texttt{results/theory\_checks.json} (\texttt{code/theory\_checks.py}, \texttt{code/t3\_check.py}).

\begin{itemize}
\item \textbf{\Cref{thm:t4}(a).} $\tau$ is a bijection of $\Z_{10}$, so the subset-independence argument applies.
\item \textbf{\Cref{thm:t4}(b).} $|\mathbb E_x\omega^{j(x-\tau(x))}|=0.2$ for odd $j$ and $0$ for even $j$. For a random code of
64 words of length 16 (first coordinate fixed, minimum distance 5), the exact correlations of all 2,016 pairs, computed
from the product formula, equal $0.05\cdot0.2^{|\Delta|}$ to within $7\times10^{-21}$. Five pairs checked by Monte Carlo
on $2\times10^6$ strings agree within 1.4 standard errors (SE $6.4\times10^{-5}$; three pairs within one SE, the others at 1.02 and
1.38 SE).
\item \textbf{\Cref{thm:t4}(c).} For a random $160$-$64$-$1$ ReLU network at initialisation, the variance over the 64
labelings of the square-loss gradient, estimated on $2\times10^5$ shared samples, is $2.3\times10^{-6}$; the bound is
$7.1\times10^{-3}$. The estimate equals the finite-sample noise floor ($2.2\times10^{-6}$), so this check shows only
that the bound is not violated; it cannot tell how small the true variance is.
\item \textbf{\Cref{thm:t3}(a)}, day-of-week MLP trained on 1900--2099 for 2,000 steps. The six unseen leading-digit
columns stay exactly at their initial values under Adam (deviation 0.0). Under AdamW they equal $(1-\eta\lambda)^TW_0$
to within $3.2\times10^{-6}$, the float32 rounding of the repeated product. Seen columns move by up to 0.17--0.18. For
Roman numerals, 100 of the numerals 1--999 contain M, which is why we do not apply (a) there.
\item \textbf{\Cref{thm:t1,thm:t2}, \cref{lem:margin}.} Checked by the certificates themselves (\cref{sec:certs},
App.~\ref{app:float}) and by the failure-point runs (\cref{tab:x3}).
\end{itemize}

\paragraph{Bayes gaps.} \Cref{tab:x7} gives the optimal-statistic tier. Hamming BER is given to five significant
digits, because the 6\,dB values differ only in the fourth. The written decoder differs from block ML on at most
0.011\% of words. Bitwise MAP has 1.83\% fewer bit errors than block ML at 0\,dB and 0.41\% fewer at 4\,dB. In the
motion rows, the written argmax equals the first maximiser of $n_{11}$ on every pair, as \cref{prop:motion} says. Both columns use the same pairs; the Bayes column is the expected accuracy with a random tie-break (ties on 4--22\% of pairs at coherence $\le0.2$), so the small gaps are tie-breaking luck; the
GD range at coherence $\ge0.5$ includes a dense-dot setting (99.46\%).

\begin{table}[!h]
\caption{Optimal-statistic tier. Hamming: BER \% on $10^6$ words per $E_b/N_0$ (rate-4/7 scaling). Motion: accuracy
\% on 20,000 fresh pairs (SE $\le0.35$ points).}
\label{tab:x7}
\centering\footnotesize
\begin{tabular}{@{}lccc@{}}
\toprule
Hamming(7,4) & written & block ML & bitwise MAP \\
\midrule
0\,dB & 8.2797 & 8.2807 & 8.1288 \\
2\,dB & 2.8874 & 2.8883 & 2.8503 \\
4\,dB & 0.53135 & 0.53145 & 0.52928 \\
6\,dB & 0.037625 & 0.037650 & 0.037475 \\
\midrule
Motion & written & Bayes & GD, 60k pairs \\
\midrule
coherence 0.2 & 88.26 & 88.21 & 78.36 \\
coherence 0.1 & 53.23 & 53.08 & 44.17 \\
$32\times32$, coherence 0.1 & 90.74 & 90.79 & 77.12 \\
coherence $\ge0.5$ (3 settings) & 100 & 100 & 99.46--100 \\
\bottomrule
\end{tabular}
\end{table}

\paragraph{Mutation testing.} For four networks we drew 200 single-parameter mutants each ($w\to0$, $w\pm1$, $-w$),
decided each mutant with its certificate, and counted how many mutants that break exactness a random test of 20,000
domain inputs misses (\texttt{code/x8b\_mutation.py}). Mutants are judged in exact arithmetic (float64 is exact here: every scaled absolute sum $M_i$ on the network's binary grid stays far below $2^{53}$). Missed: ALU 0 of 186; day of week 1 of 172; routing 29 of 158
(18\%), where one route's unit covers a small part of the address space; Luhn 12 of 51 with 20,000 uniform
strings (11 of 51 with the suite's 40,000-number Luhn test set, the protocol of the first version). Pooled under one
protocol: 42 of 567 (7.4\%). The certificates decided every mutant.



\end{document}